%% file: main.tex
\documentclass[10pt,twocolumn,letterpaper]{article}

\usepackage[pagenumbers]{cvpr} 

\usepackage{graphicx}
\usepackage{amsmath}
\usepackage{amssymb}
\usepackage{booktabs}
\usepackage{comment}
\usepackage{bm}
\usepackage{adjustbox}
\usepackage{array}
\newcolumntype{R}[2]{%
    >{\adjustbox{angle=#1,lap=1.3\width-(#2)}\bgroup}%
    l%
    <{\egroup}%
}
\usepackage{pifont}
\usepackage{stmaryrd}
\usepackage{multirow}
\usepackage{enumitem}
\usepackage{flushend}
\usepackage{makecell}
\usepackage{setspace}
\usepackage{cuted}
\usepackage{colortbl}
\usepackage{titletoc}
\usepackage[dvipsnames]{xcolor}

\usepackage[accsupp]{axessibility} 

\newcommand{\parag}[1]{\smallskip\noindent\textbf{#1}\enspace}

\usepackage[pagebackref,breaklinks,colorlinks]{hyperref}

\usepackage[capitalize]{cleveref}
\crefname{section}{Sec.}{Secs.}
\Crefname{section}{Section}{Sections}
\Crefname{table}{Table}{Tables}
\crefname{table}{Tab.}{Tabs.}

\def\cvprPaperID{***} 
\def\confName{CVPR}
\def\confYear{2023}

\hypersetup{
    colorlinks=true,
    linkcolor=red,
    citecolor=blue,
    urlcolor=magenta,
    breaklinks=true
}

\begin{document}
\title{RangeViT: Towards Vision Transformers \\ for 3D Semantic Segmentation in Autonomous Driving}

\author{
Angelika Ando$^{1,2,}$\thanks{This project was done during an internship at Valeo.ai.} ,
Spyros Gidaris$^{1}$,
Andrei Bursuc$^{1}$,
Gilles Puy$^{1}$,
Alexandre Boulch$^{1}$,
Renaud Marlet$^{1,3}$ 
\vspace*{10pt} \\
$^{1}$Valeo.ai, Paris, France \,\,\,\,\,\,\,\,\, $^{2}$Centre for Robotics, Mines Paris, Universit{\'e} PSL, Paris, France \\
$^{3}$LIGM, Ecole des Ponts, Univ.\ Gustave Eiffel, CNRS, Marne-la-Vall{\'e}e, France
}

\maketitle

\begin{abstract}
\vspace{-5pt}
    Casting semantic segmentation of outdoor LiDAR point clouds as a 2D problem, e.g., via range projection, is an effective and popular approach. These projection-based methods usually benefit from fast computations and, when combined with techniques which use other point cloud representations, achieve state-of-the-art results. Today, projection-based methods leverage 2D CNNs but recent advances in computer vision show that vision transformers (ViTs) have achieved state-of-the-art results in many image-based benchmarks. In this work, we question if projection-based methods for 3D semantic segmentation can benefit from these latest improvements on ViTs. We answer positively but only after combining them with three key ingredients: (a) ViTs are notoriously hard to train and require a lot of training data to learn powerful representations. By preserving the same backbone architecture as for RGB images, we can exploit the knowledge from long training on large image collections that are much cheaper to acquire and annotate than point clouds. We reach our best results with pre-trained ViTs on large image datasets. (b) We compensate ViTs' lack of inductive bias by substituting a tailored convolutional stem for the classical linear embedding layer. (c) We refine pixel-wise predictions with a convolutional decoder and a skip connection from the convolutional stem to combine low-level but fine-grained features of the the convolutional stem with the high-level but coarse predictions of the ViT encoder. With these ingredients, we show that our method, called RangeViT, outperforms existing projection-based methods on nuScenes and SemanticKITTI. The code is available at {{\tt \small \url{https://github.com/valeoai/rangevit}}}.
\end{abstract}


\begin{figure}[!t]
    \centering
    \includegraphics[trim={0cm 0cm 0cm 0cm},width=\linewidth]{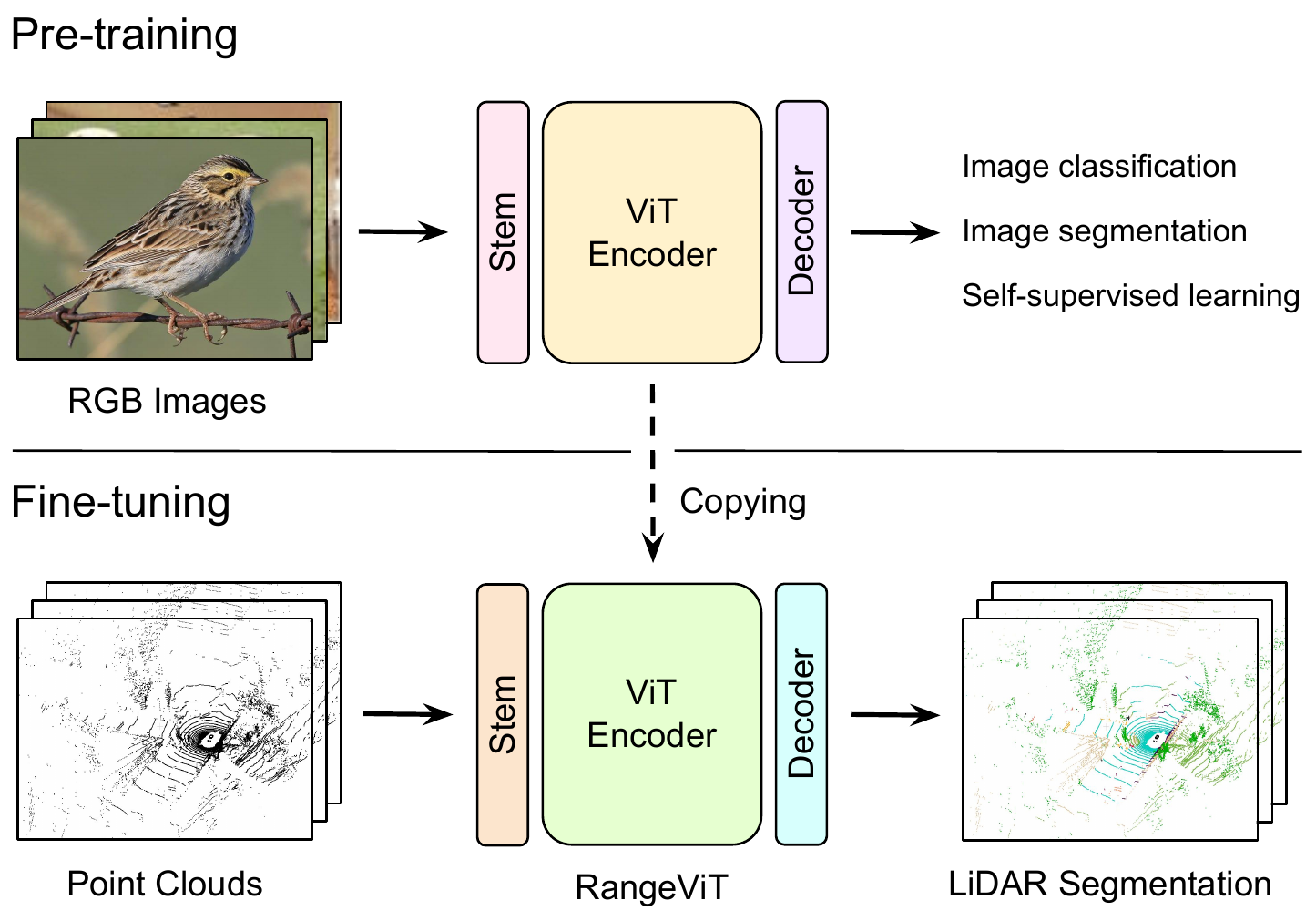}
\caption{
\textbf{Exploiting vision transformer (ViT) architectures and weights for LiDAR point cloud semantic segmentation.} We leverage the flexibility of transformer-based architectures to re-purpose them with minimal changes for processing sparse point clouds in autonomous driving tasks. The common ViT backbone across modalities allows to effectively transfer weights pre-trained on large image repositories towards improving point cloud segmentation performance with fine-tuning.
}
\label{fig:training_strategy}
\vspace{-8pt}
\end{figure}

\input{sections/introduction.tex}
\input{sections/related_work.tex}

\input{sections/approach.tex}
\input{sections/experiments.tex}
\input{sections/conclusion.tex}

\vspace*{-10pt}

\paragraph{{\footnotesize
Acknowledgements.}}
{\footnotesize
We would like to express our gratitude to Matthieu Cord and Hugo Touvron for their insightful comments, Robin Strudel \etal~\cite{strudel2021segmenter} for kindly providing ViT-S pre-trained on Cityscapes, and Oriane Sim{\'e}oni for her valuable contribution to certain experiments.
This work was performed using HPC resources from GENCI-IDRIS (Grants 2022-AD011012884R1, 2022-AD011013413 and 2022-AD011013701).
The authors acknowledge the support of the French Agence Nationale de la Recherche (ANR), under grant ANR-21-CE23-0032 (project MultiTrans).
}

{\small
\bibliographystyle{ieee_fullname}
\bibliography{egbib}
}

\renewcommand{\thesection}{\Alph{section}}
\appendix
\input{sections/supplementary_material.tex}

\end{document}

%% file: sections/introduction.tex
\section{Introduction} \label{sec:introduction}

Semantic segmentation of LiDAR point clouds permits vehicles to perceive their surrounding 3D environment independently of the lighting condition, providing useful information to build safe and reliable vehicles. A common approach to segment large scale LiDAR point clouds is to project the points on a 2D surface 
and then to use regular CNNs, originally designed for images, to process the projected point clouds~\cite{aksoy2020salsanet, cortinhal2020salsanext, milioto2019rangenet++, zhang2020polarnet, kochanov2020kprnet, xu2020squeezesegv3}. 
Recently, Vision Transformers (ViTs) were introduced as an alternative to convolutional neural networks for processing images~\cite{dosovitskiy2020image}: images are divided into patches which are linearly embedded into a high-dimensional space to create a sequence of visual tokens; these tokens are then consumed by a pure transformer architecture~\cite{vaswani2017attention} to output deep visual representations of each token. Despite the absence of almost any domain-specific inductive bias apart from the image tokenization process, ViTs have a strong representation learning capacity~\cite{dosovitskiy2020image} and achieve excellent results on various image perception tasks, such as image classification~\cite{dosovitskiy2020image}, object detection~\cite{carion2020end} or semantic segmentation~\cite{strudel2021segmenter}. 

Inspired by this success of ViTs for image understanding, we propose to implement projection-based LiDAR semantic segmentation with a pure vision transformer architecture at its core. Our goals are threefold in doing so: 
(1) Exploit the strong representation learning capacity of vision transformer 
for LiDAR semantic segmentation; 
(2) Work towards unifying network architectures used for processing LiDAR point clouds or images so that any advance in one domain benefits to both; 
(3) Show that one can leverage ViTs pre-trained on large-size natural image datasets for LiDAR point cloud segmentation. 
The last goal is crucial because the downside of having few inductive biases in ViTs is that they underperform when trained from scratch on small or medium-size datasets and that, for now, the only well-performing pre-trained ViTs ~\cite{dosovitskiy2020image,strudel2021segmenter,caron2021emerging} publicly available are trained on large collections of images that can be acquired, annotated and stored easier than point clouds.

In this context, our main contribution is a ViT-based LiDAR segmentation approach that compensates ViTs' lack of inductive biases on our data and that achieves state-of-the-art results among projection-based methods. To the best of our knowledge, although works using ViT architectures on dense indoor point clouds already exists~\cite{yu2022point, zhao2021point}, this is the first solution using ViTs for the LiDAR point clouds of autonomous driving datasets, which are significantly sparser and noisier than the dense depth-map-based points clouds found in indoor datasets. Our solution, RangeViT, starts with a classical range projection to obtain a 2D representation of the point cloud~\cite{cortinhal2020salsanext, milioto2019rangenet++, kochanov2020kprnet, xu2020squeezesegv3}. Then, we extract patch-based visual tokens from this 2D map and feed them to a plain ViT encoder~\cite{dosovitskiy2020image} to get deep patch representations. These representations are decoded using a lightweight network to obtain pixel-wise label predictions, which are projected back to the 3D point cloud.

Our finding is that this ViT architecture needs three key ingredients to reach its peak performance. First, we leverage ViT models pre-trained on large natural image datasets for LiDAR segmentation and demonstrate that our method benefits from them despite the fact that natural images display little resemblance with range-projection images. Second, we further compensate for ViTs' lack of inductive bias by substituting the classical linear embedding layer with a multi-layer convolutional stem. Finally, we refine pixel-wise predictions with a convolutional decoder and a skip connection from the convolutional stem to combine low-level but fine-grain features of the convolutional stem with the high-level but coarse predictions of the ViT encoder.

In summary, our contributions are the following:
\textbf{(1)} To the best of our knowledge, we are the first to exploit the strong representation learning capacity of vision transformers architectures for 3D semantic segmentation from LiDAR point clouds. By revisiting in the context of our problem the tokenization process of the ViT's encoder and adding a light-weight convolutional decoder for refining the coarse patch-wise ViT representations, we derive a simple but effective projection-based LiDAR segmentation approach, which we call RangeViT.
\textbf{(2)} Furthermore, as shown in \cref{fig:training_strategy}, the proposed approach allows someone to harness ViT models pre-trained on the RGB image domain for the LiDAR segmentation problem. Indeed, despite the large gap between the two domains, we empirically demonstrate that using such pre-training strategies improves segmentation performance.
\textbf{(3)} Finally, our RangeViT approach, despite its simplicity, achieves state-of-the-art results among projection-based segmentation methods.

%% file: sections/related_work.tex
\section{Related work} \label{sec:related}

\subsection{CNNs for Point Cloud Segmentation}

\parag{2D Methods.}
Several works~\cite{behley2019semantickitti, cortinhal2020salsanext, kochanov2020kprnet, landrieu2018large, milioto2019rangenet++, hu2020randla, xu2020squeezesegv3, zhang2020deep, zhang2020polarnet, zhuang2021perception, razani2021lite, thomas2019kpconv, triess2020scan} project the 3D point cloud into the 2D space with range, perspective or bird’s-eye-view (BEV) projection and process the projected images with 2D CNNs. For instance, PolarNet~\cite{zhang2020polarnet} employs bird’s-eye-view projection to polar coordinates and then processes the bird-eye-view features with a Ring CNN. DarkNetSeg~\cite{behley2019semantickitti}, SalsaNext~\cite{cortinhal2020salsanext}, KPRNet~\cite{kochanov2020kprnet}, RangeNet++~\cite{milioto2019rangenet++}, Lite-HDSeg~\cite{razani2021lite} and SqueezeSegV3~\cite{xu2020squeezesegv3} use range projection and then process the input images with a U-Net-like architecture. PMF~\cite{zhuang2021perception}, a multi-modal segmentation approach for point clouds and RGB images~\cite{bai2022transfusion, li2022deepfusion, wang2021pointaugmenting, zhang2022cat, zhuang2021perception}, projects the 3D points onto 2D camera frames and processes them together with the RGB images using a dual-stream CNN network.

\parag{3D Methods.}
Instead of the 2D space, voxel-based approaches~\cite{graham20183d, han2020occuseg, meng2019vv, tchapmi2017segcloud} process the point clouds in their 3D space by first dividing the 3D space into voxels using cartesian coordinates and then applying 3D convolutions. Cylinder3D~\cite{zhu2021cylindrical} show that dividing the 3D space into voxels using cylindrical coordinates instead of cartesian improves the segmentation performance. Although voxel-based methods consider the geometric properties of the 3D space, their drawback is that they are computationally expensive.

Also, for indoor-scene point cloud segmentation there are many methods~\cite{engelmann20203d, pham2019jsis3d, thomas2019kpconv, yan2020pointasnl, velivckovic2017graph, wang2019dynamic, wang2019graph, wu2019pointconv} relying on PointNet-inspired architectures~\cite{qi2017pointnet} that directly process raw 3D points. However, these approaches cannot be easily adapted to outdoor-scene point cloud segmentation, which we consider in this work, as the large quantity of points cause computational difficulties.

\subsection{ViTs for Point Cloud Segmentation}

The recent rise to prominence of ViT architectures in computer vision has inspired a series of works for indoor point cloud segmentation~\cite{liu2022masked, wang2022bridged, yu2022point, zhao2021point, zhou2022self}.

Processing point clouds with a ViT model can be challenging and computationally expensive due to the large number of points. Point-BERT~\cite{yu2022point} uses farthest point sampling~\cite{qi2017pointnet++} and the KNN algorithm to define input tokens for the transformer.
Also, it proposes a self-supervised pre-training strategy for point data inspired by the masked token modeling approaches in the RGB image~\cite{he2022masked} and text~\cite{devlin2018bert} domains. Point Transformer~\cite{zhao2021point} has a U-Net-like architecture but without convolutional layers and it integrates self-attention mechanism in all of its blocks. Finally, Bridged Transformer~\cite{wang2022bridged} jointly processes point clouds and RGB images with a fully transformer-based architecture. 

Using a transformer-based architecture for the semantic segmentation task on outdoor LiDAR point clouds remains challenging. To the best of our knowledge, there has been no work published yet about outdoor LiDAR point cloud semantic segmentation with a ViT-based architecture.

\subsection{Transfer Learning from Images to Point Clouds}

ViT models have the capacity to learn powerful representations, but require vast amounts of training data for it. However, large LiDAR point cloud datasets are less common due to the costly and time consuming annotation procedure. Recent works~\cite{gupta2016cross, liu2021learning, sautier2022image, xu2021image2point, qian2022pix4point, wang2022can} explore transfer learning of 2D image pre-trained models to 3D point clouds.

Image2Point~\cite{xu2021image2point} converts a 2D CNN into a 3D sparse CNN~\cite{xu2021image2point} by inflating 2D convolutions into 3D convolutions, which can be done by repeating the 2D kernel along the third dimension. SLidR~\cite{sautier2022image} is a self-supervised pre-training method on LiDAR data. It uses a student-teacher architecture, where the 2D teacher network pre-trained on images transfers information into the 3D student network. 
Pix4Point~\cite{qian2022pix4point} and Simple3D-Former~\cite{wang2022can} study transfer learning from images to indoor point clouds with fully transformer-based architectures. They adapt the tokenizer and the head layer to be specialized for 3D point cloud data. Pix4Point~\cite{qian2022pix4point} applies farthest point sampling~\cite{qi2017pointnet++}, the KNN algorithm and then a graph convolution on the aggregated neighborhoods to extract input tokens for the ViT encoder. Simple3D-Former~\cite{wang2022can} memorizes the ImageNet~\cite{deng2009imagenet} representation of 2D image classification by incorporating a KL divergence term in the loss between its 2D image classification predictions and those obtained from a fixed ImageNet~\cite{deng2009imagenet} pre-trained ViT network.

%% file: sections/approach.tex
\section{RangeViT} \label{sec:approach}

\begin{figure*}[!t]
    \centering
    \includegraphics[trim={0cm 0cm 0cm 0cm},width=0.97\linewidth]{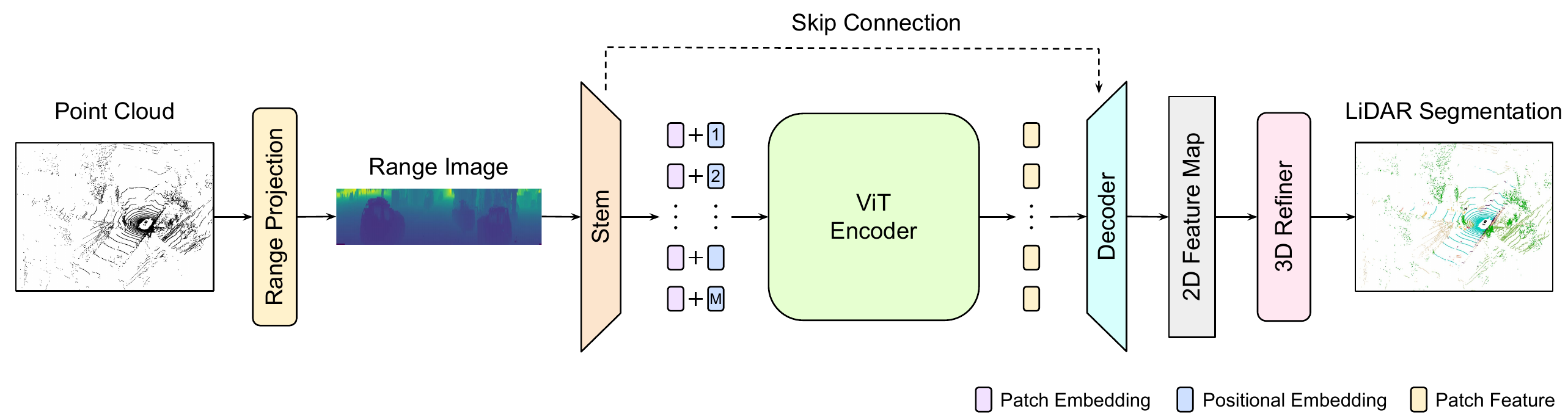}
\caption{\textbf{Overview of RangeViT architecture.} 
First, the point cloud is projected in a 2D space with range projection. Then, the produced range image is processed by the convolutional stem, the ViT encoder and the decoder to obtain a 2D feature map. It is then processed by a 3D refiner layer for 3D point-wise predictions. Note that there is a single skip connection between the convolutional stem and the decoder.}
\label{model_architecture}
\end{figure*}

In this section, we describe our ViT-based LiDAR semantic segmentation approach, for which we provide an overview in \cref{model_architecture}.

\subsection{General architecture}

We represent a LiDAR point cloud of $N$ points with a matrix $\mathbf{P} \in \mathbb{R}^{N \times 4}$. Each point $p = (x, y, z, i) \in \mathbf{P}$ has Cartesian coordinates denoted by $(x, y, z)$ and LiDAR intensity denoted by $i$, and is annotated with a label $\ell \in \{1, \ldots, K \}$ encoding one of the $K$ semantic classes.

\paragraph{Range projection.} The input of our RangeViT backbone is a 2D representation of the input point cloud. We use the well-known range projection~\cite{milioto2019rangenet++}. Each point $\mathbf{p} \in \mathbf{P}$ with coordinates $(x, y, z)$ is projected on a range image of size $H \times W$. The projected 2D coordinates satisfies
\begin{align} \label{eq:range_proj}
    \begin{pmatrix}
        h \\
        w
    \end{pmatrix}
    =
    \begin{pmatrix}
        \frac{1}{2} \left( 1 - \arctan (y, x) \pi^{-1} \right) W \\
        \left( 1 - (\arcsin (z, r^{-1}) + |f_{\text{down}}|) f_{v}^{-1} \right) H
    \end{pmatrix},
\end{align}
where $f_{v} = |f_{\text{down}}| + |f_{\text{up}}|$ is the vertical field-of-view of the LiDAR sensor. We associate $C{=}5$ low-level features $(r, x, y, z, i)$ to each projected point, where $r = \sqrt{x^2 + y^2 + z^2}$ is the range of the corresponding point (i.e., its distance from the LiDAR sensor), to create the range image $\mathbf{I} \in \mathbb{R}^{C \times H \times W}$. Note that if more than one point is projected onto the same pixel, 
then only the feature with the smallest range is kept. Pixels with no point projected on them have their features filled with zeros. 

\paragraph{Convolutional stem.} 
In a standard ViT, the image is divided into $M$ patches of size $P_H \times P_W$, which are linearly embedded to provide $D$-dimensional visual tokens. Yet, our empirical study in \cref{sec:ablation_exps} shows that this tokenization process of standard ViTs is far from optimal on both our task and datasets of interest. In order to bridge this potential domain gap between range images and standard ViTs, we replace the embedding layer with a non-linear convolutional stem~\cite{xiao2021early}. Non-linear convolutional stems have been shown to increase optimization stability and predictive performance of ViTs~\cite{xiao2021early}, whereas we leverage them primarily for \emph{steering} range images towards ViT-like inputs.

The first part of the convolutional stem consists of the first $4$ residual blocks of SalsaNext~\cite{cortinhal2020salsanext}, called context module. This context module captures short-range dependencies of the 3D points projected in the range image and produces pixel-wise features with $D_{\text{h}}$ channels at the same resolution of the input range image, hence the tensor of context features $\mathbf{t}_{\text{c}}$ has size $D_{\text{h}} \times H \times W$\footnote{The first 3 residual layers actually have $D_{\text{in}}=32$ channels, which is typically smaller than $D_{\text{h}}$.}.
Then, in order to produce tokens compatible with the input of a ViT, we use an average pooling layer that reduces the spatial dimensions of the context features $\mathbf{t}_{\text{c}}$ from $H \times W$ to $(H/P_H) \times (W/P_W)$, and use a final 1$\times$1 convolutional layer with $D$ output channels. The convolutional stem thus yields $M = (H W) / (P_H P_W)$ visual tokens $\mathbf{v}_1, \ldots, \mathbf{v}_M$ of dimension $D$, i.e., matching the input dimension and number of tokens of a traditional ViT.

\paragraph{ViT encoder.}

The output of the convolutional stem can be fed directly to a ViT~\cite{dosovitskiy2020image}. The input $\mathbf{t}_0$ to our ViT encoder is prepared by stacking all the visual tokens $\mathbf{v}_1, \ldots, \mathbf{v}_M$ and a classification token $\mathbf{v}_{\text{class}} \in \mathbb{R}^{D}$, to which we add the positional embeddings $\mathbf{E}_{\text{pos}} \in \mathbb{R}^{(M + 1) \times D}$: 
\begin{align}
    \mathbf{t}_0 = [\mathbf{v}_{\text{class}},\mathbf{v}_1, \ldots, \mathbf{v}_M] + \mathbf{E}_{\text{pos}}.
\end{align}
The input tokens are then transformed by the ViT encoder to obtain an updated sequence of tokens $\mathbf{t}_L \in \mathbb{R}^{(M + 1) \times D}$, where $L$ denotes the number of transformer blocks. Then, we remove the classification token from $\mathbf{t}_L$ to keep only the deep patch representations $\mathbf{t}'_L \in \mathbb{R}^{M \times D}$.

\paragraph{Decoder.} 
The representations $\mathbf{t}'_L$ provided by the ViT encoder are patch representations which are unfortunately too coarse to obtain good point predictions. Therefore, we use a decoder to refine these coarse patch representations. First, $\mathbf{t}'_L$ is reshaped in form of a 2D feature map of size $D \times H/P_H \times W/P_W$. Our convolutional decoder consists of a 1$\times$1 convolution layer with $D_{\text{h}} P_H P_W$ output channels, followed by a Pixel Shuffle layer~\cite{shi2016real} which yields feature maps with shape $D_{\text{h}} \times H \times W$, i.e., with the same resolution as the original range image. While the convolutional decoder can still produce coarse features or decoding artifacts, the Pixel Shuffle is particularly effective in recovering fine information from features. Then, we concatenate these features with the context features $\mathbf{t}_{\text{c}}$ from the convolutional stem and use a series of two convolutional layers with 3$\times$3 and 1$\times$1 kernels respectively, each of them followed by Leaky ReLU and batch normalisation, to obtain the refined feature map $\mathbf{t}_{\text{dec}} \in \mathbb{R}^{D_{\text{h}} \times H \times W}$.

\paragraph{3D refiner.}
Ultimately, we need to convert the pixel-wise features from the range image space into point-wise predictions in the 3D space. Most prior range-projection based methods first make pixel-wise class predictions and then un-project them to the 3D space, where at inference time often there is a post-processing step relying, e.g., on K-NN~\cite{milioto2019rangenet++} or CRFs~\cite{krahenbuhl2011efficient}. The purpose of the latter post-process step is to fix segmentation mistakes related to the projection and processing of 3D points in a 2D space (e.g., multiple 3D points being projected on the same pixel, or 2D boundary prediction errors for points that are actually far away in the 3D space). Instead, we follow the approach of KPRNet~\cite{kochanov2020kprnet} that proposes an end-to-end approach that learns this post-processing step with a KPConv~\cite{thomas2019kpconv} layer.

KPConv~\cite{thomas2019kpconv} is a point convolution technique which works directly on the original 3D points. 
It permits to leverage the underlying geometry of the 3D point clouds to refine features at the point level.
So, in our network, we project the 2D feature maps $\mathbf{t}_{\text{dec}}$ of the 2D decoder back to original 3D points by bilinear upsampling, thus obtaining point-wise features with shape $N \times D_{\text{h}}$, where $N$ is the number of 3D points. Then, these point features are given to the KPConv layer as input along with the 3D coordinates of the corresponding points, which outputs $D_{\text{h}}$-dimensional point features. Finally, the logits $\mathbf{s} \in \mathbb{R}^{N \times K}$ are obtained by applying a BatchNorm, a ReLU and a final point-wise linear layer on these point features.

\subsection{Implementation details} 

\parag{Training loss.}
We use the sum of the multi-class focal loss~\cite{lin2017focal} and the Lov{\'a}sz-softmax loss~\cite{berman2018lovasz}. The focal loss is a scaled version of the cross-entropy loss~\cite{goodfellow2016deep} adapting its penalty to the hardness of the samples, making it suited for datasets with class imbalance, such as semantic segmentation. The Lov{\'a}sz-softmax is developed specifically for semantic segmentation and built to optimize the mIoU.

\parag{Inference.} \label{sec:approach_inference}
As in~\cite{strudel2021segmenter}, we use a sliding-window method during inference. The network actually never sees the entire range images during training but only crops extracted from it. At inference, the range image is divided into overlapping crops of the same size as those used during training. The corresponding 2D features at the output of the decoder are then averaged to reconstruct the entire feature map, which is then processed by our 3D refinement layer.

%% file: sections/experiments.tex
\section{Experiments} \label{sec:experiments}

\subsection{Experimental Setup} \label{sec:experimental_setup}

\parag{Datasets and metrics.}
We validate our approach for 3D point cloud semantic segmentation on two different commonly used datasets: nuScenes~\cite{caesar2020nuscenes} and SemanticKITTI~\cite{behley2019semantickitti}. We conduct most of our ablation studies on nuScenes and compare against previous works on both nuScenes and SemanticKITTI.
As evaluation metric, we use the mean Intersection over Union (mIoU)~\cite{everingham2015pascal}.

NuScenes~\cite{caesar2020nuscenes} consists of 1,000 scenes of 20 seconds in Boston and Singapore with various urban scenarios, lighting and weather conditions. The LiDAR sensor has 32 laser beams. Furthermore, there are 16 annotated semantic classes and the dataset is split into 28,130 training and 6,019 validation point cloud scans.

SemanticKITTI~\cite{behley2019semantickitti} is created from the KITTI Vision Odometry Benchmark~\cite{geiger2012we} and consists of urban scenes collected in Germany. Sequences 00-10 are used for training, except sequence 08 which is used for validation. There are 19,130 training and 4,071 validation scans. Sequences 11-21 are used for test and they contain 20,351 scans. The LiDAR sensor has 64 laser beams and there are 19 annotated semantic classes.

\parag{Model and pre-trained weights.}
For all experiments, we use the ViT-S/16 model~\cite{dosovitskiy2020image} as the encoder. It has $L{=}12$ layers, 6 attention heads and $D{=}384$ channels, amounting to approximately 21M parameters. Unless otherwise stated, (a) this ViT encoder is initialized with weights pre-trained on ImageNet21k~\cite{deng2009imagenet} for classification and then fine-tuned on Cityscapes~\cite{cordts2016cityscapes} for semantic image segmentation~\cite{strudel2021segmenter}; (b) the stem, decoder and 3D refiner layers, which are always randomly initialized, use $D_{\text{h}}=256$ feature channels.

\begin{table}[!t]\centering
\footnotesize
\addtolength{\tabcolsep}{-2pt}
\centering
\begin{tabular}{llccc}
\toprule
\textbf{Stem} & \textbf{Decoder} & \textbf{Refiner} & \textbf{mIoU} & \textbf{\#Params}\\\midrule
Linear & Linear &            & 65.52 & 22.0M\\
Conv   & Linear &            & 69.82 & 22.8M\\
Conv   & UpConv &            & 73.83 & 24.6M\\
Conv   & UpConv & \checkmark & \textbf{74.60} & 25.2M \\
\bottomrule
\end{tabular}
\vspace{-8pt}
\caption{\textbf{Model ablations.} Results on the nuScenes validation set with $D_{\text{h}}=192$. The linear stem refers to the linear patch embedding layer. When the 3D refiner layer (Refiner column) is not used, we use the K-NN post-processing technique~\cite{milioto2019rangenet++}.} 
\label{tab:model_ablation}
\vspace{-8pt}
\end{table}

\parag{Optimization.}
We use the AdamW optimizer~\cite{loshchilov2017decoupled} with $\beta_1 = 0.9$, $\beta_2 = 0.999$ and weight decay $0.01$. The batch-size is $32$ and $16$ for nuScenes and SemanticKITTI, respectively. For the learning rate $lr$, we use a linear warm-up from $0$ to its peak value for $10$ epochs and then we decrease it over the remaining training epochs to $0$ with a cosine annealing schedule~\cite{loshchilov2016sgdr}. In SemanticKITTI, we use 60 training epochs and the peak $lr$ is $4 \times 10^{-4}$. In nuScenes, we use 150 epochs and the peak $lr$ is $8 \times 10^{-4}$ when training from Cityscapes, ImageNet21k and random initializations, and $2 \times 10^{-4}$ when training from DINO initialization.

\parag{Data augmentations.}
As point cloud augmentations we use: (a) flips over the $y$ axis (the vertical axis on the range image), (b) random translations, and (c) random rotations between $\pm 5^{\circ}$ (using the roll, pitch and yaw angles). All augmentations are applied randomly with probability $0.5$. Finally, after range projection, we take a random image crop with a fixed size of $32 \times 384$ for nuScenes and $64 \times 384$ for SemanticKITTI. Note that the full size of a range image is $32 \times 2048$ for nuScenes and $64 \times 2048$ for SemanticKITTI.

\subsection{What makes a ViT architecture for 3D semantic segmentation?} \label{sec:ablation_exps}

\begin{table}[!t]\centering
\footnotesize
\addtolength{\tabcolsep}{-2pt}
\centering
\begin{tabular}{lcccc}
\toprule
\textbf{$D_{\text{h}}$ size} & 64    & 128   & 192   & 256\\\midrule
\textbf{\#Params}            & 22.7M & 23.7M & 25.2M & 27.1M\\\midrule  
\textbf{nuScenes mIoU}       & 74.00 & 74.12 & 74.60 & \textbf{75.21}\\
\textbf{SemanticKITTI mIoU}       & 58.48 & \textbf{60.73} & 60.45    & 60.55\\
\bottomrule
\end{tabular}
\vspace{-8pt}
\caption{\textbf{Impact of the $D_{\text{h}}$ channel size.} Used in the stem, decoder, and 3D refiner layers. Results on the nuScenes and the SemanticKITTI validation sets.} 
\label{tab:impact_width_h}
\vspace{-8pt}
\end{table}

Our aim is to adapt to LiDAR point clouds with the fewest possible modifications of the standard ViT architecture~\cite{dosovitskiy2020image}, which already follows faithfully the original self-attention Transformer design~\cite{vaswani2017attention}. Differently from ViT that performs classification on images, RangeViT performs semantic segmentation of point cloud-derived range images. We study here the possible choices for the input processing and decoder layers using the nuScenes dataset.

\parag{Stem and decoder.}
In \cref{tab:model_ablation}, we study the impact of the convolutional stem and the UpConv decoder. To that end, we report results using a linear patch embedding~\cite{dosovitskiy2020image} as a stem and a linear decoder.\footnote{The linear decoder consists of a simple 1$\times$1 conv. layer, which predicts patch-wise classification scores, and a bilinear upsampling to reach the input image resolution.}
Starting from a model with linear stem and linear decoder, introducing the proposed convolutional stem leads to a significant mIoU boost ($65.52 \rightarrow 69.82$). This highlights the importance of having a non-linear network component for producing appropriate input token features for the ViT backbone. In addition, when using image pre-trained ViTs (\cref{sec:pretraining_exps}) the convolutional stem can effectively steer the distribution of input range images towards the image-based feature distribution the ViT has been pre-trained on, leading to smoother fine-tuning. Then, also replacing the linear decoder with the proposed UpConv decoder leads to another notable improvement on mIoU ($69.8 \rightarrow 73.83$), validating this design choice. Finally, replacing the K-NN refinement \cite{milioto2019rangenet++} with the KPConv layer (in the 3D refiner layer), leads to a small but non-trivial mIoU increase ($73.83 \rightarrow 74.60$).

In \cref{tab:impact_width_h}, we study what is the impact, on the mIoU, of changing the number of feature channels $D_{\text{h}}$ on the stem, decoder and 3D refiner layer. On nuScenes, we observe that the results gradually increase when increasing the dimension $D_{\text{h}}$. In the SemanticKITTI case, $D_{\text{h}} = 128$ gives the best results and further increasing $D_{\text{h}}$ does not help.
Note that the size of the ViT's embeddings is fixed at $D=384$, regardless of the $D_{\text{h}}$ size.

\begin{table}[!t]\centering
\footnotesize
\addtolength{\tabcolsep}{-2pt}
\resizebox{1.\linewidth}{!}{
\centering
\begin{tabular}{lccccccc}
\toprule
\textbf{Patch size} 
& $16 \times 16$ 
& $8 \times 8$ 
& $4 \times 16$ 
& $4 \times 8$ 
& $4 \times 4$ 
& $2 \times 16$ 
& $2 \times 8$\\
\midrule
\textbf{mIoU} 
& 68.45 
& 72.04 
& 72.72 
& 73.30 
& 73.70 
& 73.88 
& \textbf{75.21}\\
\midrule
\textbf{\#Tokens}
& 49 
& 193 
& 193 
& 385 
& 769 
& 385 
& 769\\
\textbf{Train time} 
& $\times$1 
& $\times$1.02 
& $\times$1.02 
& $\times$1.13 
& $\times$1.43 
& $\times$1.13 
& $\times$1.43\\
\bottomrule
\end{tabular}
}
\vspace{-8pt}
\caption{\textbf{Patch size ablations.} Results on the nuScenes validation set. The mIoU scores are in increasing order. The train time is relative to the 16$\times$16 patch size case.} 
\label{tab:patch_size}
\vspace{-8pt}
\end{table}

\parag{What patch-size for range image ``tokenization''?}
The size of the input patch tokens is an essential factor for controlling the trade-off between speed and accuracy performance without changing the number of parameters in the model. Intuitively, reducing the patch-size results in a finer representation, but also a longer input sequence that takes longer to process. Conversely, larger patches lead to a coarser representation, yet faster inference. The performance impact of the patch-size can sometimes match the one of model size~\cite{strudel2021segmenter}. 
For simplicity, standard ViT models use square patches of size 32, 16 or 8, as they are typically trained on square images. In contrast, range images have a much different aspect ratio (high width) and different layout of the content in the image (rows of points corresponding to LiDAR beams). With this insight, we revisit the practice of using square patches and look into rectangular patches with different aspect ratio (1:2, 1:4, 1:8) that would better capture the specific local patterns of the range images.

\cref{tab:patch_size} shows the mIoU scores for different patch sizes for dividing the range image to patch tokens. Note that the convolutional stem first produces pixel-wise features and then, given a patch-size $(P_H, P_W)$, reduces the spatial dimensions of these features to $(H/P_H, W/P_W)$ using local average pooling. We observe that smaller patch area is better but not necessarily with a square patch, commonly used for ViT models. The wide range images benefit more from rectangular, patches ($2 \times 8$ performing best). The smaller patches enable the extraction of more fine-grained information leading to more precise predictions for smaller and thinner objects and points at object boundaries.

\subsection{Exploiting image pre-trained ViTs}
\label{sec:pretraining_exps}
\begin{table}[!t]\centering
\footnotesize
\addtolength{\tabcolsep}{-2pt}
\centering
\begin{tabular}{lcccc}
\toprule
\textbf{Pre-training} & Rand & DINO & IN21k & CS\\\midrule
\textbf{mIoU}         & 72.37 & 73.33 & 74.77 & \textbf{75.21}\\
\bottomrule
\end{tabular}
\vspace{-8pt}
\caption{\textbf{ViT pre-training on RGB images.} Comparison of different weight initializations of the ViT encoder on the nuScenes validation set. Rand: randomly initialized. The positional embeddings are initialized with the corresponding pre-trained weights or randomly when training from scratch. The convolutional stem, the decoder and the 3D refiner layer are randomly initialized.} 
\label{tab:pretraining}
\vspace{-8pt}
\end{table}

\begin{figure}[!t]
    \centering
    \includegraphics[trim={0cm 0cm 0cm 0cm},width=\linewidth]{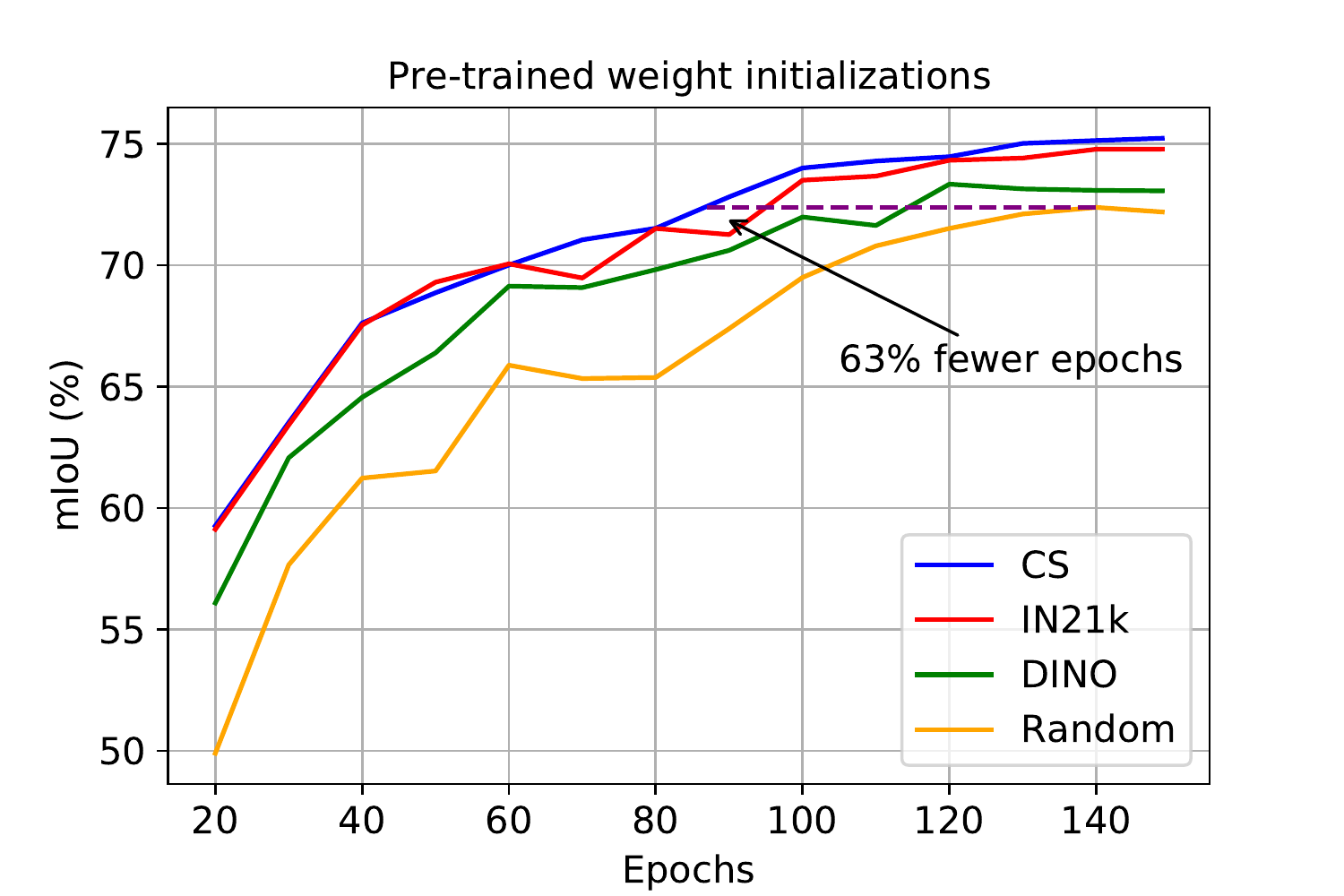}
\vspace{-15pt}
\caption{\textbf{Training efficiency  with image-pretrained ViTs.}
Comparison of the validation mIoU curves for different weight initializations of the ViT encoder on the nuScenes validation set.}
\label{fig:pre_trained_weights}
\vspace{-8pt}
\end{figure}

So far, we have studied the architectural choices necessary for using ViT models for point cloud semantic segmentation. The final RangeViT model preserves the ViT backbone intact allowing us to initialize it with weights from models pre-trained on large datasets. We study now whether using such an initialization helps or not and what fine-tuning strategies would be more suitable in this context.

\parag{Is pre-training on RGB images beneficial?}
In \cref{tab:pretraining}, we study the effect of transferring to our task ViT models pre-trained on natural RGB images.
In particular, we explore initializing RangeViT's backbone with ViTs pre-trained:
(a) on supervised ImageNet21k classification~\cite{dosovitskiy2020image} (entry IN21k),
(b) on supervised image segmentation on Cityscapes with Segmenter~\cite{strudel2021segmenter} (entry CS), which in its turn was pre-trained with IN21k, and 
(c) with the DINO~\cite{caron2021emerging} self-supervised approach on ImageNet1k (entry DINO).

We observe that, despite the large domain gap, using ViT models pre-trained on RGB images is always better than training from scratch on LiDAR data (entry Rand). For instance, using the IN21k and CS pre-trained ViTs leads to improving the mIoU scores by $2.4$ and $2.8$ points, respectively. Additionally, as we see in \cref{fig:pre_trained_weights}, which plots the nuScenes validation mIoU as a function of the training epochs, using such pre-trained ViTs leads to faster training convergence of  the LiDAR segmentation model. We argue that this is a highly interesting finding. It means that our RangeViT approach, by being able to use off-the-shelf pre-trained ViT models, can directly benefit from current and future advances on the training ViT models with natural RGB images, a very active and rapidly growing research field~\cite{touvron2021going, touvron2022three, he2022masked, radford2021learning}. Furthermore, from the small difference between the mIoU scores with the IN21k and CS pre-trainings, we infer that the pre-training can lead to consistent performance improvements even if it is not on the strongly-supervised image segmentation task, which requires expensive-to-annotate datasets.

\begin{table}[!t]\centering
\footnotesize
\addtolength{\tabcolsep}{-2pt}
\centering
\begin{tabular}{c|ccc|cc}
\toprule
& \multicolumn{3}{c|}{Fine-tuning} & IN21k & CS \\
Model & LN         & ATTN       & FFN        & \multicolumn{2}{c}{mIoU}\\
\midrule
(a) & \checkmark & \checkmark & \checkmark & 74.79 & 75.21 \\
\midrule
(b) &            &            &            & 67.88 & 68.03 \\
(c) & \checkmark &            &            & 69.08 & 69.31 \\
(d) & \checkmark & \checkmark &            & 73.56 & 72.77 \\
(e) & \checkmark &            & \checkmark & 75.11 & 75.47 \\
\bottomrule
\end{tabular}
\vspace{-8pt}
\caption{\textbf{Partial ViT fine-tuning.}
We use a ViT encoder pre-trained on ImageNet21k (IN21k) or Cityscapes (CS) and partially fine-tune RangeViT on the nuScenes training set.
The convolutional stem, the positional embeddings and the UpConv decoder are always fine-tuned.
LN: fine-tuning the LayerNorm layers of the ViT encoder.
ATTN: fine-tuning the multi-head attention layers of the ViT encoder.
FFN: fine-tuning the feed-forward network layers of the ViT encoder.
Model (a) is full fine-tuning of the network.
The results are reported on the nuScenes validation set.
} \label{tab:finetuning}
\end{table}

\begin{table}[!t]\centering
\footnotesize
\addtolength{\tabcolsep}{-2pt}
\centering
\begin{tabular}{l|cc|cc|c}
\toprule
\textbf{Encoder} & ViT-S$^\dagger$ & ViT-S & RN50$^\dagger$ & RN50 & Identity\\\midrule
\textbf{mIoU (\%)}        & 67.88 & \textbf{74.77}  & 60.48 & 72.30 & 53.73\\
\bottomrule
\end{tabular}
\vspace{-8pt}
\caption{\textbf{Ablating encoder backbones.}
ViT-S and RN50 are pre-trained on IN21k.
$\dagger$: the encoder remains frozen during training.} 
\label{tab:ablate_encoders}
\vspace{-8pt}
\end{table}

\parag{Which ViT layers is better to fine-tune?}
In the image domain, several practices have emerged for fine-tuning a pre-trained convolutional network on a downstream dataset. Taking into consideration the domain gap between the pre-training and downstream data and the amount of labeled data available, the entire network can be fine-tuned or only a part of it. Here the domain gap between RGB images and range images is major and we would expect a full fine-tuning of the network to be better. To understand how much prior knowledge of the image pre-trained ViT is useful for point clouds, we study different fine-tuning strategies: fine-tuning all layers, only the attention (ATTN) layers or only the feedforward network (FFN) layers.

In \cref{tab:finetuning}, we show results with these different fine-tuning strategies. 
Interestingly, the best result are not achieved with full fine-tuning (model (a)) but when the attention layers are kept frozen (model (e)). This suggests that the pre-trained ATTN layers are already well learnt and ready to generalize to range images. With CS pre-training ATTN layers may capture the layout of the scenes more easily without much fine-tuning, as the LiDAR scans have been also acquired from urban road scenes. Fine-tuning FFN may have more impact due to the different specifics of the LiDAR data compared to RGB images. In addition, FFN layers are in practice easier and more stable to optimize (they are essentially fully-connected and normalization layers) than ATTN layers that usually require more careful hyper-parameter selection. These findings align with recent ones from the image domain~\cite{touvron2022three}, confirming that the convolutional stems do steer the input LiDAR data to behave like image data once on in the ViT backbone.

\begin{figure}[!t]
    \centering
    \includegraphics[trim={0cm 0cm 0cm 0cm},width=\linewidth]{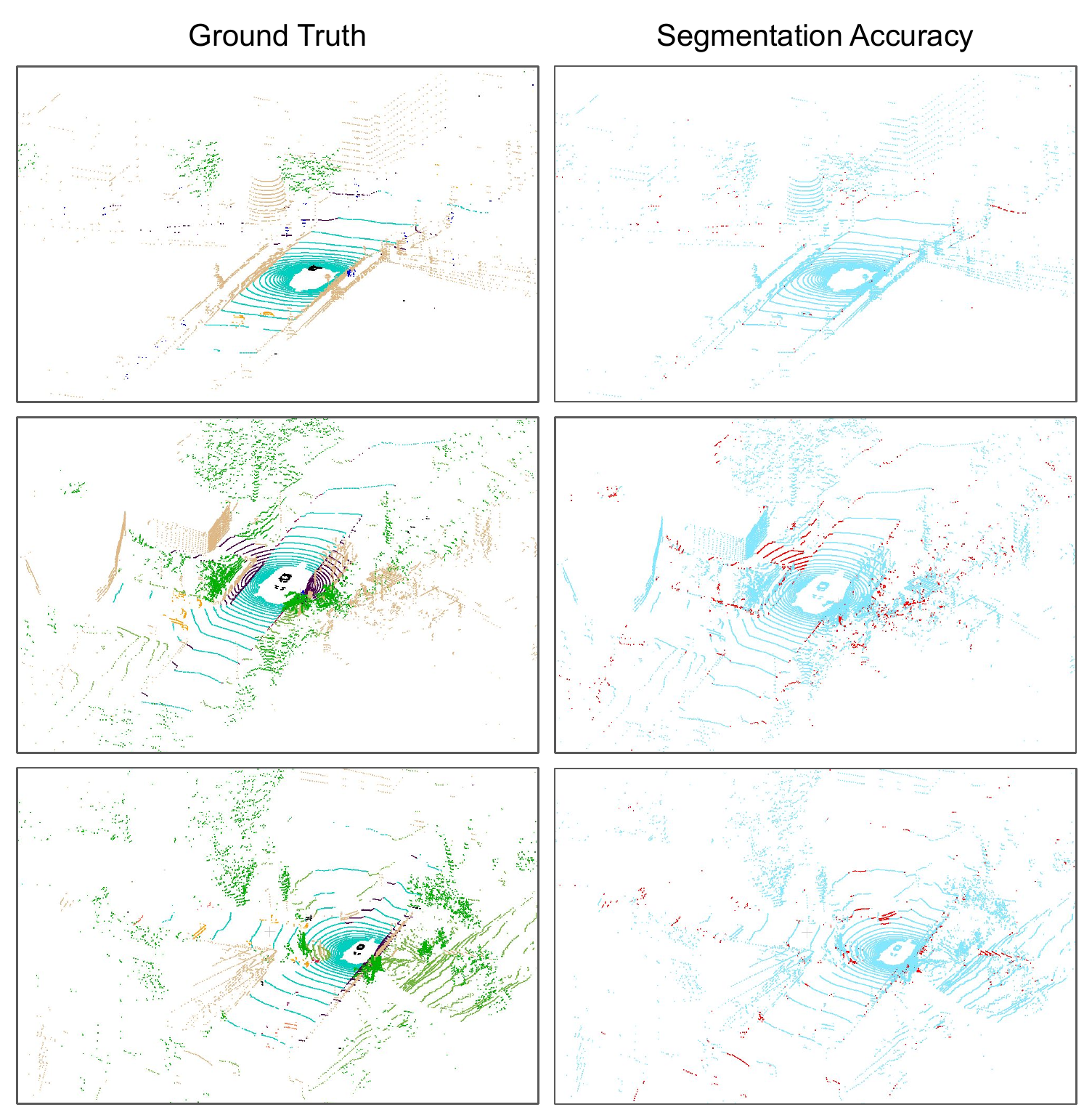}
    \vspace{-20pt}
\caption{\textbf{Visualizing the segmentation accuracy.} Left: ground truth segmentation of validation point clouds of nuScenes, coloured based on their label. Right: segmentation accuracy of RangeViT -- good predictions are in blue and the bad ones in red.}
\label{nuscenes_qual_res}
\end{figure}

\begin{table*}[!t]
\centering
\footnotesize
\addtolength{\tabcolsep}{-0.5pt}
\resizebox{1.\linewidth}{!}{
\centering
\begin{tabular}{lccccccccccccccccc}
\toprule
\textbf{Method} 
&\rotatebox{90}{barrier} 
&\rotatebox{90}{bicycle} 
&\rotatebox{90}{bus} 
&\rotatebox{90}{car} 
&\rotatebox{90}{construction} 
&\rotatebox{90}{motorcycle} 
&\rotatebox{90}{pedestrian} 
&\rotatebox{90}{traffic cone} 
&\rotatebox{90}{trailer} 
&\rotatebox{90}{truck} 
&\rotatebox{90}{driveable} 
&\rotatebox{90}{other flat} 
&\rotatebox{90}{sidewalk} 
&\rotatebox{90}{terrain} 
&\rotatebox{90}{manmade} 
&\rotatebox{90}{vegetation} 
&\rotatebox{90}{\textbf{mIoU (\%)}} \\
\midrule
\multicolumn{2}{l}{\textbf{Voxel-based}}\\
\: Cylinder3D~\cite{zhu2021cylindrical} &\textbf{76.4} &\textbf{\textcolor{blue}{40.3}} &\textbf{91.3} &\textbf{93.8} &\textbf{51.3} &78.0 &\textbf{78.9} &64.9 &62.1 &\textbf{84.4} &\textbf{96.8} &\textbf{71.6} &\textbf{76.4} &\textbf{75.4} &\textbf{90.5} &\textbf{87.4} &\textbf{76.1} \\
\midrule
\multicolumn{2}{l}{\textbf{2D Projection-based}}\\
\: RangeNet++~\cite{milioto2019rangenet++} &66.0 &21.3 &77.2 &80.9 &30.2 &66.8 &69.6 &52.1 &54.2 &72.3 &94.1 &66.6 &63.5 &70.1 &83.1 &79.8 &65.5 \\
\: PolarNet~\cite{zhang2020polarnet} &74.7 &28.2 &85.3 &\textbf{\textcolor{blue}{90.9}} &35.1 &77.5 &71.3 &58.8 &57.4 &76.1 &\textbf{\textcolor{blue}{96.5}} &71.1 &\textbf{\textcolor{blue}{74.7}} &\textbf{\textcolor{blue}{74.0}} &87.3 &85.7 &71.0 \\
\: SalsaNext~\cite{cortinhal2020salsanext} &74.8 &34.1 &85.9 &88.4 &42.2 &72.4 &72.2 &63.1 &61.3 &76.5 &96.0 &70.8 &71.2 &71.5 &86.7 &84.4 &72.2 \\
\: RangeViT-IN21k (ours) & 75.1 & 39.0 & \textbf{\textcolor{blue}{90.2}} & 88.4 & 48.0 & \textbf{\textcolor{blue}{79.2}} & \textbf{\textcolor{blue}{77.2}} & \textbf{66.4} & \textbf{\textcolor{blue}{65.1}} & 76.7 & 96.3 & 71.1 & 73.7 & 73.9 & 88.9 & 87.1 & 74.8 \\
\: RangeViT-CS (ours) & \textbf{\textcolor{blue}{75.5}} & \textbf{40.7} & 88.3 & 90.1 & \textbf{\textcolor{blue}{49.3}} & \textbf{79.3} & \textbf{\textcolor{blue}{77.2}} & \textbf{\textcolor{blue}{66.3}} & \textbf{65.2} & \textbf{\textcolor{blue}{80.0}} & 96.4 & \textbf{\textcolor{blue}{71.4}} & 73.8 & 73.8 & \textbf{\textcolor{blue}{89.9}} & \textbf{\textcolor{blue}{87.2}} & \textbf{\textcolor{blue}{75.2}} \\
\bottomrule
\end{tabular}
}
\vspace{-8pt}
\caption{\textbf{nuScenes validation set comparison} with state-of-the-art methods. The best results are bold and the second best results are blue.}
\label{tab:nus_comp_results}
\vspace{-5pt}
\end{table*}

\begin{table*}[!t]
\centering
\footnotesize
\addtolength{\tabcolsep}{-2pt}
\resizebox{1.\linewidth}{!}{
\centering
\begin{tabular}{lcccccccccccccccccccc}
\toprule
\textbf{Method} 
&\rotatebox{90}{car} 
&\rotatebox{90}{bicycle} 
&\rotatebox{90}{motorcycle} 
&\rotatebox{90}{truck} 
&\rotatebox{90}{other vehicle} 
&\rotatebox{90}{person} 
&\rotatebox{90}{bicyclist} 
&\rotatebox{90}{motorcyclist} 
&\rotatebox{90}{road} 
&\rotatebox{90}{parking} 
&\rotatebox{90}{sidewalk} 
&\rotatebox{90}{other ground} 
&\rotatebox{90}{building} 
&\rotatebox{90}{fence} 
&\rotatebox{90}{vegetation} 
&\rotatebox{90}{trunk} 
&\rotatebox{90}{terrain} 
&\rotatebox{90}{pole} 
&\rotatebox{90}{traffic sign} 
&\rotatebox{90}{\textbf{mIoU (\%)}} \\
\midrule
\multicolumn{2}{l}{\textbf{Voxel-based}}\\
\: Cylinder3D~\cite{zhu2021cylindrical} &\textbf{97.1} &\textbf{67.6} &\textbf{64.0} &\textbf{59.0} &\textbf{58.6} &\textbf{73.9} &\textbf{\textcolor{blue}{67.9}} &36.0 &91.4 &65.1 &75.5 &\textbf{\textcolor{blue}{32.3}} &91.0 &66.5 &\textbf{\textcolor{blue}{85.4}} &\textbf{71.8} &\textbf{\textcolor{blue}{68.5}} &\textbf{62.6} &\textbf{\textcolor{blue}{65.6}} &\textbf{67.8} \\
\midrule
\multicolumn{2}{l}{\textbf{2D Projection-based}}\\
\: RangeNet++~\cite{milioto2019rangenet++} &91.4 &25.7 &34.4 &25.7 &23.0 &38.3 &38.8 &4.8 &91.8 &65.0 &75.2 &27.8 &87.4 &58.6 &80.5 &55.1 &64.6 &47.9 &55.9 &52.2 \\
\: PolarNet~\cite{zhang2020polarnet} &93.8 &40.3 &30.1 &22.9 &28.5 &43.2 &40.2 &5.6 &90.8 &61.7 &74.4 &21.7 &90.0 &61.3 &84.0 &65.5 &67.8 &51.8 &57.5 &54.3 \\
\: SqueezeSegV3~\cite{xu2020squeezesegv3} &92.5 &38.7 &36.5 &29.6 &33.0 &45.6 &46.2 &20.1 &91.7 &63.4 &74.8 &26.4 &89.0 &59.4 &82.0 &58.7 &65.4 &49.6 &58.9 &55.9 \\
\: SalsaNext~\cite{cortinhal2020salsanext} &91.9 &48.3 &38.6 &\textbf{\textcolor{blue}{38.9}} &31.9 &60.2 &59.0 &19.4 &91.7 &63.7 &75.8 &29.1 &90.2 &64.2 &81.8 &63.6 &66.5 &54.3 &62.1 &59.5 \\ 
\: KPRNet~\cite{kochanov2020kprnet} &\textbf{\textcolor{blue}{95.5}} &54.1 &47.9 &23.6 &\textbf{\textcolor{blue}{42.6}} &\textbf{\textcolor{blue}{65.9}} &65.0 &16.5 &\textbf{93.2} &\textbf{73.9} &\textbf{80.6} &30.2 &\textbf{\textcolor{blue}{91.7}} &\textbf{\textcolor{blue}{68.4}} &\textbf{85.7} &69.8 &\textbf{71.2} &58.7 &64.1 &63.1 \\
\: Lite-HDSeg~\cite{razani2021lite} &92.3 &40.0 &\textbf{\textcolor{blue}{55.4}} &37.7 &39.6 &59.2 &\textbf{71.6} &\textbf{54.1} &93.0 &68.2 &78.3 &29.3 
&91.5 &65.0 &78.2 &65.8 &65.1 &59.5 &\textbf{67.7} &63.8 \\
\: RangeViT-CS (ours) & 95.4 & \textbf{\textcolor{blue}{55.8}} & 43.5 & 29.8 & 42.1 & 63.9 & 58.2 & \textbf{\textcolor{blue}{38.1}} & \textbf{\textcolor{blue}{93.1}}  & \textbf{\textcolor{blue}{70.2}} & \textbf{\textcolor{blue}{80.0}} & \textbf{32.5} & \textbf{92.0} & \textbf{69.0} & 85.3 & \textbf{\textcolor{blue}{70.6}} & \textbf{71.2} & \textbf{\textcolor{blue}{60.8}} & 64.7 & \textbf{\textcolor{blue}{64.0}} \\
\bottomrule
\end{tabular}
}
\vspace{-8pt}
\caption{\textbf{SemanticKITTI test set comparison} with state-of-the-art methods. The best results are bold and the second best results are blue.}
\label{tab:kitti_comp_results}
\vspace{-8pt}
\end{table*}

\parag{Ablating encoder backbones.}
In \cref{tab:ablate_encoders}, we replace the ViT-S encoder backbone of RangeViT with a ResNet-50 (RN50) encoder or the identity function (i.e., the decoder follows directly the stem). Both ViT-S and RN50 are pre-trained on IN21k. We see that switching from ViT-S to RN50 decreases the mIoU from $74.77\%$ to $72.30\%$. Furthermore, when the backbones remain frozen (ViT-S$^\dagger$ and RN50$^\dagger$ models), we reach $67.88\%$ with ViT-S and $60.48\%$ with RN50, demonstrating that ViT features are more appropriate for transfer learning to LiDAR data than CNNs. Finally, with the Identity backbone, we achieve $53.73\%$, which is more than 14 points worse than the ViT-S$^\dagger$ model that has the same number of learnable parameters.

\subsection{Comparison to the state-of-the-art}

\cref{tab:nus_comp_results,tab:kitti_comp_results} report the final comparison on the nuScenes validation set and on the SemanticKITTI test set including class-wise IoU scores. We observe that our model achieves superior mIoU performance compared to prior 2D projection based methods on both datasets, reducing the gap with the strong voxel-based Cylinder3D~\cite{zhu2021cylindrical} method.

\parag{Class-wise IoU result analysis.}
In \cref{tab:nus_comp_results,tab:kitti_comp_results}, we can see that RangeViT often achieves the best or the second best class-wise IoU scores. In both datasets, the classes are imbalanced and due to the sparsity and varying density of LiDAR point clouds, some classes (e.g., bicycle, pedestrian) are represented with few, not necessarily structured points per scene, so it is difficult recognize them. This problem could possibly be reduced by jointly processing point clouds and RGB images, 
since RGB images might provide extra cues about the outline and the shape of these objects.

\subsection{Qualitative Results}

We visualize the 3D point clouds with the ground truth semantic labels as well as the predictions with CloudCompare~\cite{girardeau2016cloudcompare}. \cref{nuscenes_qual_res} shows the predictions of RangeViT on three validation point clouds of nuScenes. We notice minor errors such as man-made objects predicted as sidewalk and imperfect borders for the vegetation. We also remark difficulties in recognizing pedestrians and confusion between the driveable surface and the sidewalk for few points.

%% file: sections/conclusion.tex
\section{Conclusion} \label{sec:conclusion}

We studied the feasibility of leveraging (pre-trained) ViTs for LiDAR 3D semantic segmentation with projection-based methods. We discover that in spite of the significant domain gap between RGB images and range images and their high requirements of training data, ViTs can be successfully used without any changes in the original transformer backbone. We achieve this thanks to an adapted tokenization and pre-processing for the ViT encoder and a simple convolutional decoder. We show that ViTs pre-trained on large image datasets can be effectively repurposed for LiDAR segmentation towards reaching state-of-the-art performance among 2D projection methods. We release the code for our implementation and hope that it could be used as a testbed for evaluating the ability of ViT image ``foundation'' models~\cite{bommasani2021opportunities} to generalize on different domains.

\parag{Future work.}
Although the results are promising, there is still room for improvement. For instance, we identified the tokenization of LiDAR data as a crucial factor for success. As future work, we could further improve this process, e.g., with FlexiViT~\cite{beyer2022flexivit} (random patch sizes) or Perceiver IO~\cite{jaegle2021perceiver} (learning to extract tokens), and consider tokenizing raw 3D data instead of the 2D projections.

%% file: sections/supplementary_material.tex

\begin{figure*}[!t]
    \centering
    \includegraphics[trim={0cm 0cm 0cm 0cm},width=0.9\linewidth]{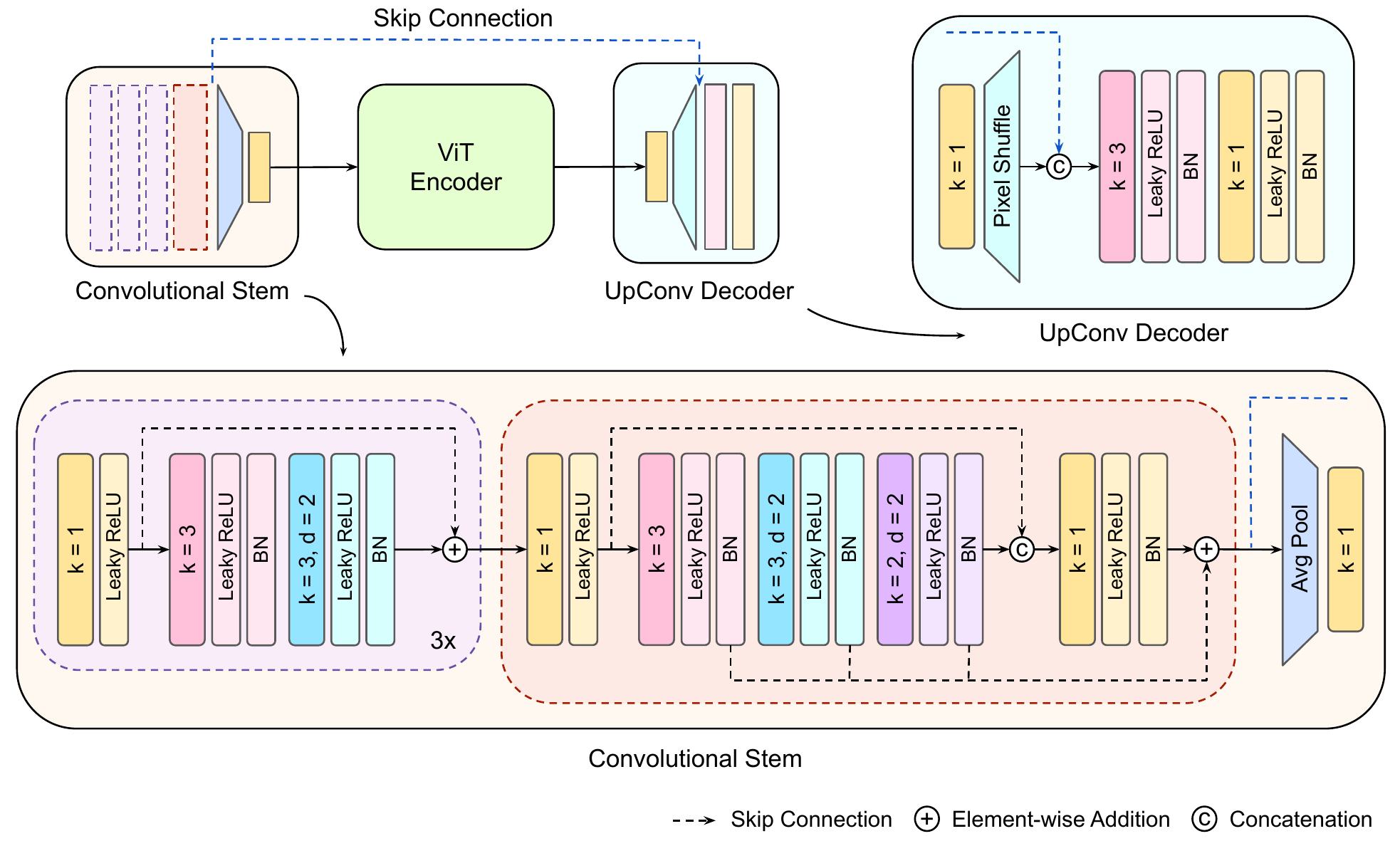}
\vspace*{-3pt}
\caption{
\textbf{Convolutional stem and UpConv decoder architecture.} In the upper left corner, there is an overview of the RangeViT architecture from the convolutional stem until the UpConv decoder. The stem and the decoder are also shown in more details. A convolution with kernel size $(s,s)$ is denoted by $k = s$ and the dilation is denoted by $d$, where it is applied. 
Note that all the rectangles in the figure with $k$ inside them are convolutional layers.
The batch normalisation layers are denoted by BN.
}
\label{fig:stem_decoder}
\end{figure*}

\section{Additional visualizations}

\begin{figure*}[!t]
    \centering
    \includegraphics[trim={0cm 0cm 0cm 0cm},width=\linewidth]{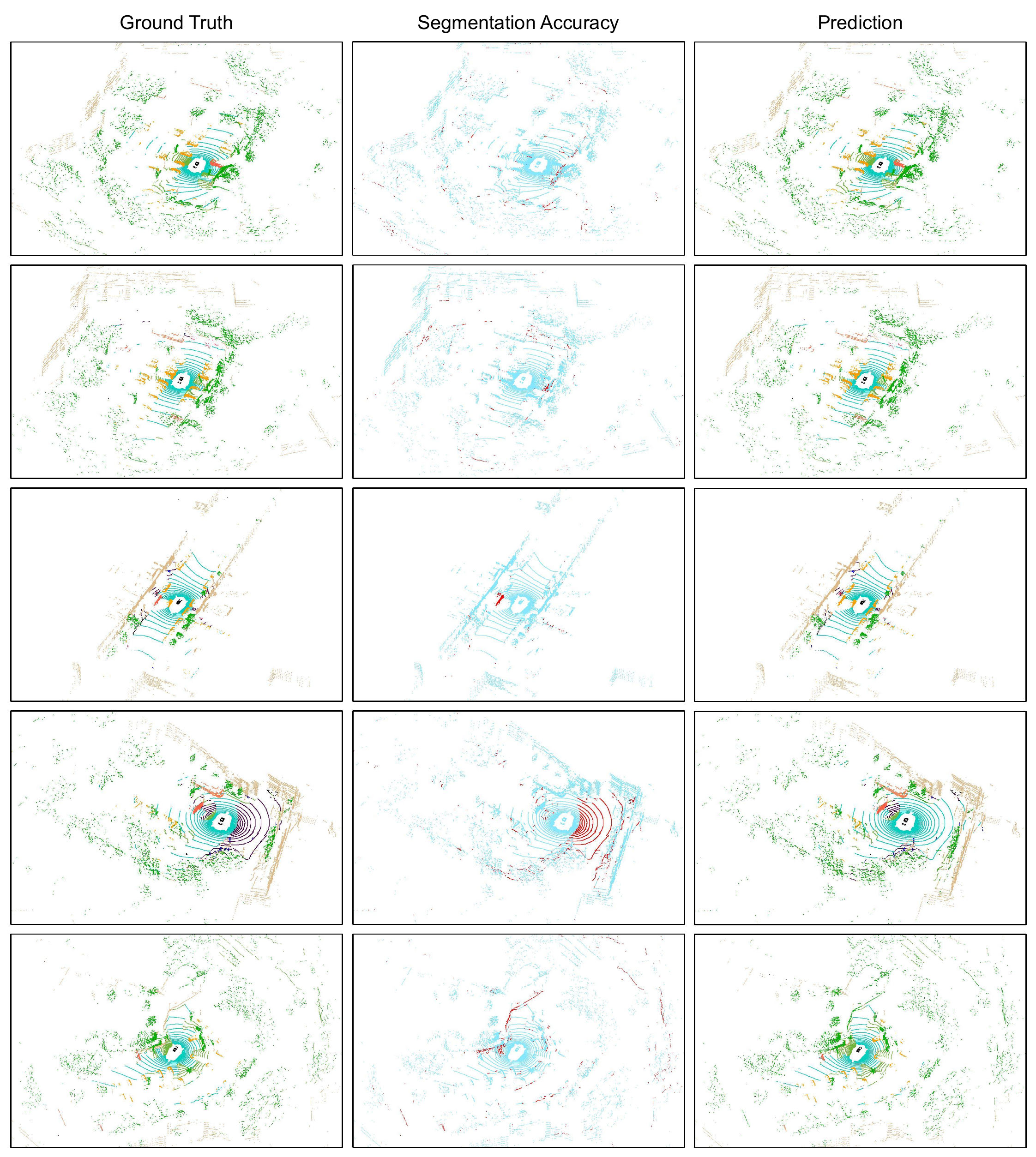}
\caption{
\textbf{Segmentation accuracy visualizations} of RangeViT on validation point clouds of nuScenes. In the left column, the points are coloured with their ground truth label colours and in the right column they are coloured with the colour of the predicted label colours. In the middle column, the good predictions are colored in blue and the bad predictions are colored in red.
}
\label{fig:vis1}
\end{figure*}

\begin{figure*}[!t]
    \centering
    \includegraphics[trim={0cm 0cm 0cm 0cm},width=\linewidth]{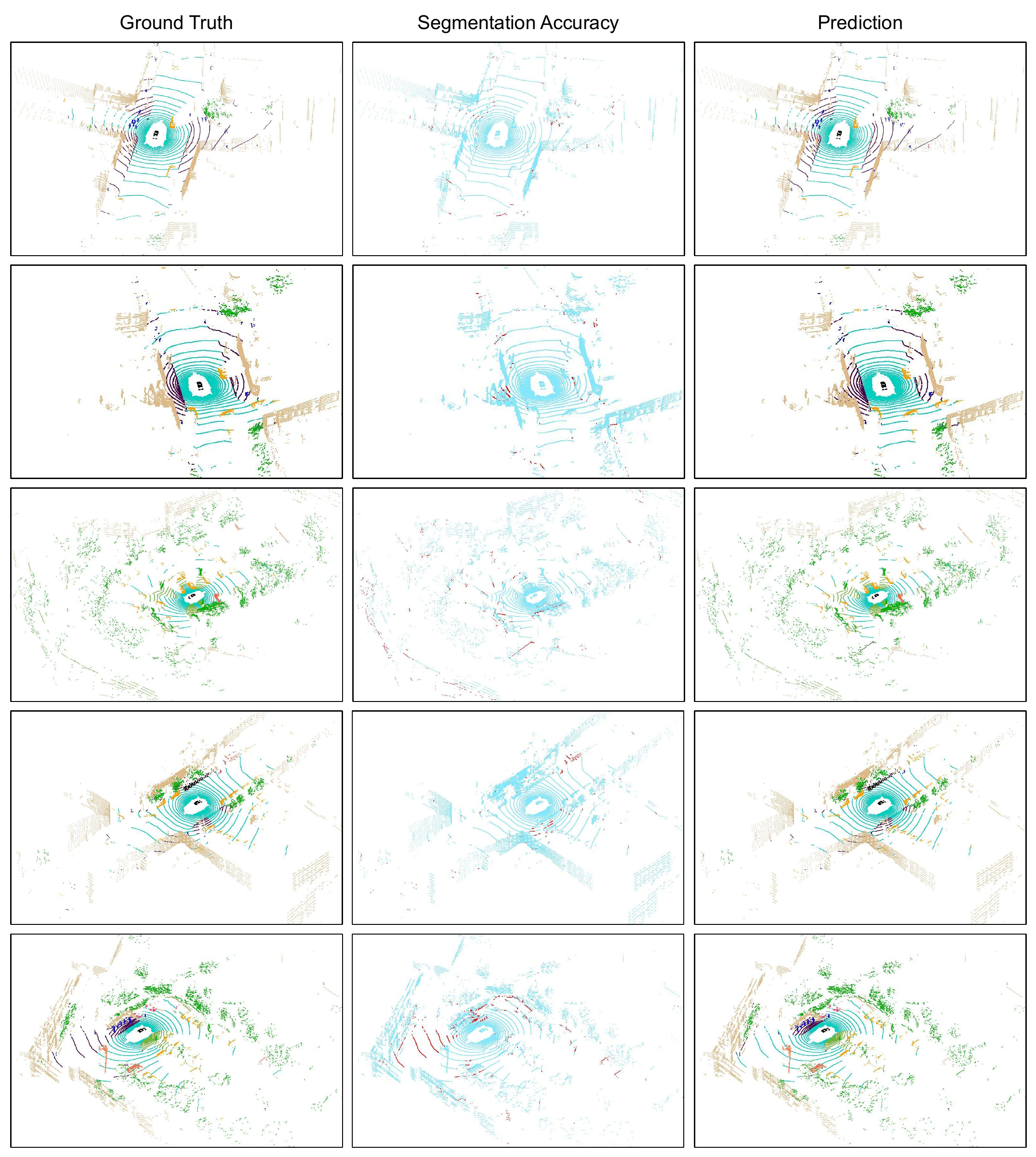}
\caption{
\textbf{Segmentation accuracy visualizations} of RangeViT on validation point clouds of nuScenes. In the left column, the points are coloured with their ground truth label colours and in the right column they are coloured with the colour of the predicted label colours. In the middle column, the good predictions are colored in blue and the bad predictions are colored in red. 
}
\label{fig:vis2}
\end{figure*}

\begin{figure*}[!t]
    \centering
    \includegraphics[trim={0cm 0cm 0cm 0cm},width=\linewidth]{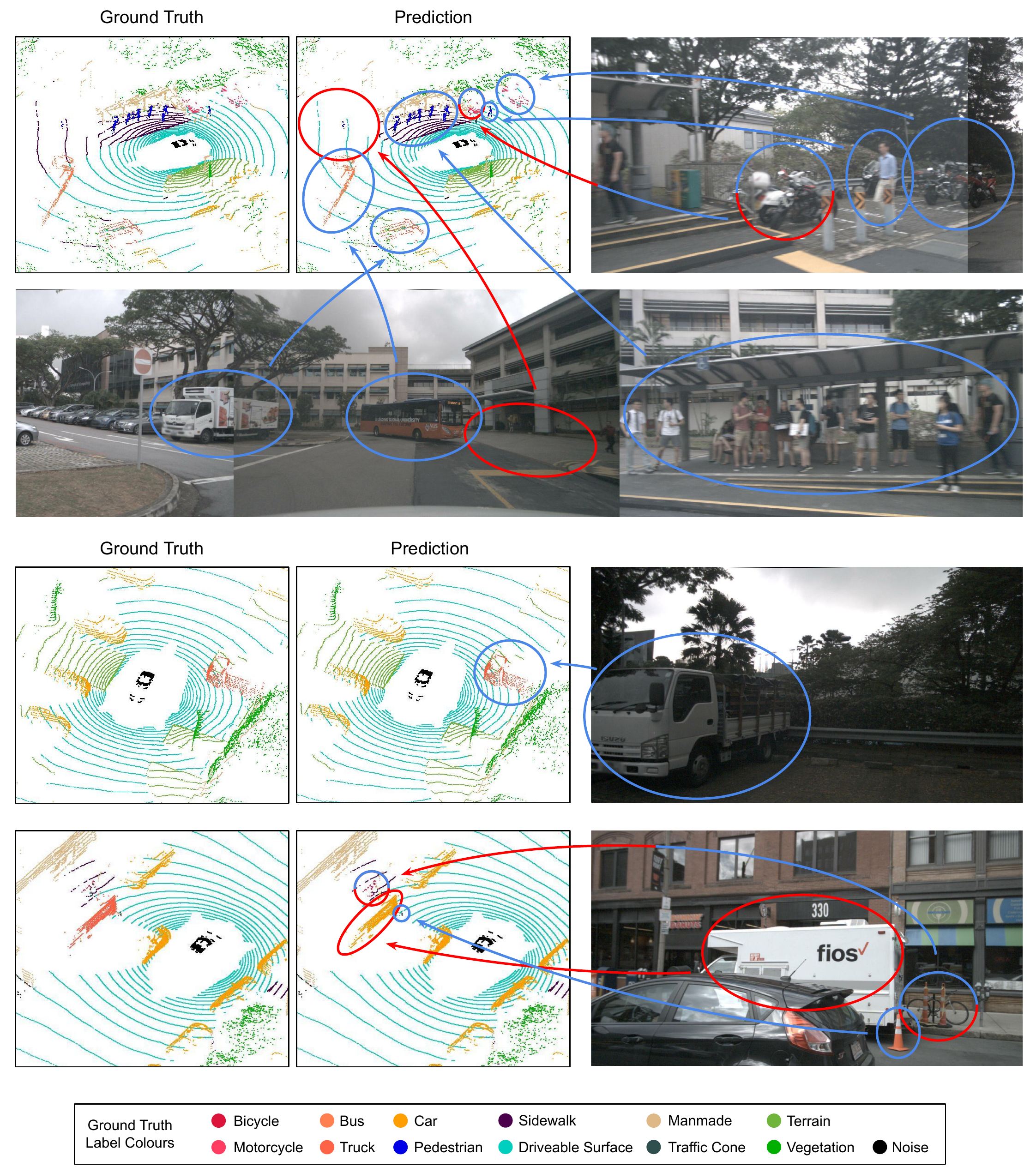}
\caption{
\textbf{Segmentation accuracy visualizations} of RangeViT on validation point clouds of nuScenes. The correct predictions are circled in blue, the incorrect ones in red and predictions that are mostly correct except for a few points are circled in half blue-half red. In the first example, the truck, the bus and the pedestrians are correctly predicted. The motorcycles are mostly correctly predicted except for a few points that were predicted as vegetation or manmade. However, part of the sidewalk is predicted as driveable surface and as we can see in the image, there is no height difference between the two, so it is difficult to recognize the sidewalk. In the second example, the truck was correctly predicted. In the third example, the truck was incorrectly predicted as a car and the traffic cones next to the bicycle were predicted as bicycle. The traffic cone next to the truck was however correctly predicted as well as the bicycle.
}
\label{fig:vis3}
\end{figure*}

\cref{fig:vis1,fig:vis2} show visualizations of the segmentation accuracy of RangeViT and \cref{fig:vis3} shows instances of correct and incorrect predictions. The visualizations are made on nuScenes validation point clouds. More information is provided in the captions of these figures.

\section{Model parameter count analysis}

In \cref{tab:model_ablation} of the main paper, we studied the impact of the non-linear convolutional stem and the UpConv decoder. In \cref{tab:model_ablation_v2}, we complete the results of \cref{tab:model_ablation} with a model (e) which, like model (a), has a linear stem and a linear decoder but for which the ViT backbone contains $L=14$ transformer layers\footnote{Note that, as pre-trained weights for the additional $2$ transformer layers of model (e), which were placed on top of the existing $12$ transformer layers of ViT-S, we used the pre-trained weights from the last available transformer layer.} instead of $L=12$. Hence, this additional model (e) and our full RangeViT solution (model (d)) have a similar number of parameters. This experiment shows that the significant performance improvement of our full solution (d) is not simply due to a higher number of parameters, since model (e) performs much worse than our RangeViT (d). Besides, our full RangeViT solution with $D_{\text{h}}=64$ (model (f)) also reaches a significantly better mIoU than models (a), (b) and (e) while having nearly the same number of parameters as model (a). This confirms that the proposed convolutional stem and UpConv decoder play an important role in the performance improvement.

\section{Computation cost comparison}

In \cref{tab:param_count}, we compare the number of parameters and the inference time of RangeViT with other LiDAR segmentation methods. RangeViT has $27.1$M parameters, which is four times more than SalsaNext~\cite{cortinhal2020salsanext} ($6.73$M), half of Cylinder3D~\cite{zhu2021cylindrical} ($55.9$M) and eight times less than KPRNet~\cite{kochanov2020kprnet} ($213.2$M). The inference time on nuScenes using the same GeForce RTX 2080 GPU is: $25$\,ms for RangeViT, $28$\,ms for SalsaNext with K-NN post-processing ($15$\,ms without it), and $49$\,ms for Cylinder3D (using the simplified and faster re-implementation of \cite{sautier2022image}).

\begin{table}[!t]\centering
\footnotesize
\addtolength{\tabcolsep}{-2pt}
\centering
\begin{tabular}{cllccccc}
\toprule
& \textbf{Stem} & \textbf{Decoder} & \textbf{Refiner} & $\pmb{L}$ & $\pmb{D_{\text{h}}}$ & \textbf{mIoU} & \textbf{\#Params}\\\midrule
(a) & Linear & Linear &            & 12 & N/A & 65.52 & 22.0M\\
(b) & Conv   & Linear &            & 12 & 192 & 69.82 & 22.8M\\
(c) & Conv   & UpConv &            & 12 & 192 & 73.83 & 24.6M\\
(d) & Conv   & UpConv & \checkmark & 12 & 192 & \textbf{74.60} & 25.2M\\\midrule
(e) & Linear & Linear &            & 14 & 192 & 65.52 & 25.6M\\\midrule
(f) & Conv   & UpConv & \checkmark & 12 & 64 & 74.00 & 22.7M\\
\bottomrule
\end{tabular}
\vspace{-8pt}
\caption{\textbf{Model ablations.} Results on the nuScenes validation set. The linear stem refers to the linear patch embedding layer. When the 3D refiner layer (Refiner column) is not used, we use the K-NN post-processing technique~\cite{milioto2019rangenet++}.} 
\label{tab:model_ablation_v2}
\end{table}

\begin{table}[!t]\centering
\footnotesize
\addtolength{\tabcolsep}{-2pt}
\centering
\begin{tabular}{lrr}
\toprule
 \textbf{Method} & \textbf{\#Params} & \textbf{Inference time}
 \\
\midrule  
SalsaNext~\cite{cortinhal2020salsanext} & 6.7M & 28\,ms\\
KPRNet~\cite{kochanov2020kprnet}        & 213.2M  &  -\\
Cylinder3D~\cite{zhu2021cylindrical}    & 55.9M & 49\,ms\\
RangeViT (ours) & 27.1M & 25\,ms\\
\bottomrule
\end{tabular}
\vspace{-8pt}
\caption{\textbf{Parameter count and inference time.} Results on the nuScenes validation set. 
} 
\label{tab:param_count}
\end{table}

\begin{table}[!t]\centering
\footnotesize
\addtolength{\tabcolsep}{-2pt}
\centering
\begin{tabular}{lccc}
\toprule
\textbf{Crop size} & $32 \times 256$ & $32 \times 384$ & $32 \times 512$ \\
\midrule
\textbf{mIoU} & 74.40 & \textbf{75.21} & 74.51 \\
\bottomrule
\end{tabular}
\vspace{-8pt}
\caption{\textbf{Impact of crop size.}
Effect of the training crop size on the nuScenes validation set with Cityscapes pre-training.
} 
\label{tab:crop_size}
\vspace{-10pt}
\end{table}

\section{Additional ablation analysis}

\parag{Impact of crop size.}
During training, we take a fixed-sized random crop from the range image, which is the input of the model. This design choice avoids computing self-attention on the whole range image in the ViT encoder, but it lacks the global information carried by the whole image. Nevertheless, our method is still successful for the semantic segmentation task since the $H \times 384$ window crop covers the whole vertical field-of-view (FOV) and one fifth of the horizontal (azimuthal) FOV ($67.5$ degrees). This is what a single nuScenes camera captures and where objects are already well identifiable. Moreover, we recall that sliding windows are also successfully used for 2D semantic segmentation with ViTs, e.g., in Segmenter~\cite{strudel2021segmenter}. 

In \cref{tab:crop_size}, we experiment with different crop sizes. As we see, the crop size has a small impact on the performance. It is also likely that tuning the learning rate and the number of epochs for these new crop sizes will reduce these small gaps even further.

\parag{Role of the classification token.}
The classification token interacts with the patch embeddings in the ViT encoder, but it is removed from the encoder output. As we do not use it directly for the semantic segmentation task, we explored its role by omitting it completely from the pipeline. Thus omitting the class token makes the mIoU drop from $75.21\%$ to $74.64\%$ and from $72.37\%$ to $72.24\%$ for the Cityscapes pre-training and no pre-training (random initialization), respectively. We hypothesize that the larger mIoU drop for Cityscapes pre-training is because, during 2D segmentation pre-training, the class token learned to carry global information that the patch tokens exploit via self-attention in order to extract better features for the segmentation task. In any case, the differences are small and thus feature extraction can still be achieved without adding the class token.

\section{Additional implementation details}

\parag{Convolutional stem and UpConv decoder.}
\cref{fig:stem_decoder} shows the detailed architecture of the convolutional stem and the UpConv decoder. All convolutions that are applied before the average pooling layer in the convolutional stem or after the Pixel Shuffle layer in the decoder do not change the spatial dimensions of the feature map $H \times W$, so appropriate paddings were applied where necessary.

As described in \cref{sec:approach} of the main paper, the average pooling layer reduces the spatial dimensions from $H \times W$ to $(H/P_H) \times (W/P_W)$. To achieve this, we use kernel size $(P_H + 1) \times (P_W + 1)$, kernel stride $P_H \times P_W$ and padding $(P_H / 2) \times (P_W / 2)$.

In the stem, the first three residual blocks use $32$ feature channels and the change of dimensions from $C=5$ input channels to $32$ channels happens in the first convolutional layer of the first residual block. The fourth residual block in the stem uses $D_h$ feature channels and similarly the change of dimensions from $32$ input channels to $D_h$ channels happens again in the first convolutional layer. Finally, the last convolutional layer in the stem (that is after the average pooling layer) uses $D$ output feature channels.

\parag{KPConv layer of the 3D Refiner.}
The KPConv~\cite{thomas2019kpconv} layer of the 3D Refiner has $D_{\text{h}}$ input and output feature channels. Its kernel size is $15$ points and the influence radius of each kernel point is $1.2$.

\parag{Replacing ViT-S with ResNet-50.}
In \cref{tab:ablate_encoders} of the main paper, we report the results that we achieve with our model when we replace the ViT-S encoder backbone with a ResNet-50 (RN50) backbone. To implement this RN50-based model, instead of the ViT-S we use the four residual blocks of RN50, to which we introduced dilations to maintain a constant spatial resolution as in ViTs. The stem and decoder remain the same, except for the number of output channels of the stem ($64$) and the input channels of the decoder ($2048$) to make them compatible with RN50. The resulting model has comparable FLOPs (the inference time for both is $25$\,ms) but much more parameters (25.2M vs 35.3M).

\parag{Range-projection for the SemanticKITTI experiments.}
To generate the 2D range images for the SemanticKITTI experiments, instead of using the spherical projection described by \cref{eq:range_proj} of the main paper, we follow \cite{triess2020scan, kochanov2020kprnet} and unfold the LiDAR scans according to the order in which they are captured by the sensor.

%% file: main.bbl
\begin{thebibliography}{10}\itemsep=-1pt

\bibitem{aksoy2020salsanet}
Eren~Erdal Aksoy, Saimir Baci, and Selcuk Cavdar.
\newblock {SalsaNet}: Fast road and vehicle segmentation in {LiDAR} point
  clouds for autonomous driving.
\newblock In {\em IV}, 2020.

\bibitem{bai2022transfusion}
Xuyang Bai, Zeyu Hu, Xinge Zhu, Qingqiu Huang, Yilun Chen, Hongbo Fu, and
  Chiew-Lan Tai.
\newblock {TransFusion}: Robust {LiDAR}-camera fusion for {3D} object detection
  with transformers.
\newblock In {\em CVPR}, 2022.

\bibitem{behley2019semantickitti}
Jens Behley, Martin Garbade, Andres Milioto, Jan Quenzel, Sven Behnke, Cyrill
  Stachniss, and Jurgen Gall.
\newblock {SemanticKITTI}: A dataset for semantic scene understanding of
  {LiDAR} sequences.
\newblock In {\em ICCV}, 2019.

\bibitem{berman2018lovasz}
Maxim Berman, Amal~Rannen Triki, and Matthew~B Blaschko.
\newblock The lov{\'a}sz-softmax loss: A tractable surrogate for the
  optimization of the intersection-over-union measure in neural networks.
\newblock In {\em CVPR}, 2018.

\bibitem{beyer2022flexivit}
Lucas Beyer, Pavel Izmailov, Alexander Kolesnikov, Mathilde Caron, Simon
  Kornblith, Xiaohua Zhai, Matthias Minderer, Michael Tschannen, Ibrahim
  Alabdulmohsin, and Filip Pavetic.
\newblock {FlexiViT}: One model for all patch sizes.
\newblock In {\em CVPR}, 2023.

\bibitem{bommasani2021opportunities}
Rishi Bommasani, Drew~A Hudson, Ehsan Adeli, Russ Altman, Simran Arora, Sydney
  von Arx, Michael~S Bernstein, Jeannette Bohg, Antoine Bosselut, Emma
  Brunskill, et~al.
\newblock On the opportunities and risks of foundation models.
\newblock In {\em arXiv}, 2021.

\bibitem{caesar2020nuscenes}
Holger Caesar, Varun Bankiti, Alex~H Lang, Sourabh Vora, Venice~Erin Liong,
  Qiang Xu, Anush Krishnan, Yu Pan, Giancarlo Baldan, and Oscar Beijbom.
\newblock {nuScenes}: A multimodal dataset for autonomous driving.
\newblock In {\em CVPR}, 2020.

\bibitem{carion2020end}
Nicolas Carion, Francisco Massa, Gabriel Synnaeve, Nicolas Usunier, Alexander
  Kirillov, and Sergey Zagoruyko.
\newblock End-to-end object detection with transformers.
\newblock In {\em ECCV}, 2020.

\bibitem{caron2021emerging}
Mathilde Caron, Hugo Touvron, Ishan Misra, Herv{\'e} J{\'e}gou, Julien Mairal,
  Piotr Bojanowski, and Armand Joulin.
\newblock Emerging properties in self-supervised vision transformers.
\newblock In {\em ICCV}, 2021.

\bibitem{cordts2016cityscapes}
Marius Cordts, Mohamed Omran, Sebastian Ramos, Timo Rehfeld, Markus Enzweiler,
  Rodrigo Benenson, Uwe Franke, Stefan Roth, and Bernt Schiele.
\newblock The {Cityscapes} dataset for semantic urban scene understanding.
\newblock In {\em CVPR}, 2016.

\bibitem{cortinhal2020salsanext}
Tiago Cortinhal, George Tzelepis, and Eren Erdal~Aksoy.
\newblock {SalsaNext}: Fast, uncertainty-aware semantic segmentation of {LiDAR}
  point clouds.
\newblock In {\em ISVC}, 2020.

\bibitem{deng2009imagenet}
Jia Deng, Wei Dong, Richard Socher, Li-Jia Li, Kai Li, and Li Fei-Fei.
\newblock {ImageNet}: A large-scale hierarchical image database.
\newblock In {\em CVPR}, 2009.

\bibitem{devlin2018bert}
Jacob Devlin, Ming-Wei Chang, Kenton Lee, and Kristina Toutanova.
\newblock {BERT}: Pre-training of deep bidirectional transformers for language
  understanding.
\newblock In {\em arXiv}, 2018.

\bibitem{dosovitskiy2020image}
Alexey Dosovitskiy, Lucas Beyer, Alexander Kolesnikov, Dirk Weissenborn,
  Xiaohua Zhai, Thomas Unterthiner, Mostafa Dehghani, Matthias Minderer, Georg
  Heigold, Sylvain Gelly, et~al.
\newblock An image is worth 16x16 words: Transformers for image recognition at
  scale.
\newblock In {\em ICLR}, 2021.

\bibitem{engelmann20203d}
Francis Engelmann, Martin Bokeloh, Alireza Fathi, Bastian Leibe, and Matthias
  Nie{\ss}ner.
\newblock {3D-MPA}: Multi-proposal aggregation for {3D} semantic instance
  segmentation.
\newblock In {\em CVPR}, 2020.

\bibitem{everingham2015pascal}
Mark Everingham, SM Eslami, Luc Van~Gool, Christopher~KI Williams, John Winn,
  and Andrew Zisserman.
\newblock The pascal visual object classes challenge: A retrospective.
\newblock In {\em IJCV}, 2015.

\bibitem{geiger2012we}
Andreas Geiger, Philip Lenz, and Raquel Urtasun.
\newblock Are we ready for autonomous driving? {The} {KITTI} vision benchmark
  suite.
\newblock In {\em CVPR}, 2012.

\bibitem{girardeau2016cloudcompare}
Daniel Girardeau-Montaut.
\newblock {CloudCompare}.
\newblock {\em France: EDF R\&D Telecom ParisTech}, 2016.

\bibitem{goodfellow2016deep}
Ian Goodfellow, Yoshua Bengio, and Aaron Courville.
\newblock {\em Deep learning}.
\newblock MIT press, 2016.

\bibitem{graham20183d}
Benjamin Graham, Martin Engelcke, and Laurens Van Der~Maaten.
\newblock {3D} semantic segmentation with submanifold sparse convolutional
  networks.
\newblock In {\em CVPR}, 2018.

\bibitem{gupta2016cross}
Saurabh Gupta, Judy Hoffman, and Jitendra Malik.
\newblock Cross modal distillation for supervision transfer.
\newblock In {\em CVPR}, 2016.

\bibitem{han2020occuseg}
Lei Han, Tian Zheng, Lan Xu, and Lu Fang.
\newblock {OccuSeg}: Occupancy-aware {3D} instance segmentation.
\newblock In {\em CVPR}, 2020.

\bibitem{he2022masked}
Kaiming He, Xinlei Chen, Saining Xie, Yanghao Li, Piotr Doll{\'a}r, and Ross
  Girshick.
\newblock Masked autoencoders are scalable vision learners.
\newblock In {\em CVPR}, 2022.

\bibitem{hu2020randla}
Qingyong Hu, Bo Yang, Linhai Xie, Stefano Rosa, Yulan Guo, Zhihua Wang, Niki
  Trigoni, and Andrew Markham.
\newblock {RandLA-Net}: Efficient semantic segmentation of large-scale point
  clouds.
\newblock In {\em CVPR}, 2020.

\bibitem{jaegle2021perceiver}
Andrew Jaegle, Sebastian Borgeaud, Jean-Baptiste Alayrac, Carl Doersch, Catalin
  Ionescu, David Ding, Skanda Koppula, Daniel Zoran, Andrew Brock, Evan
  Shelhamer, et~al.
\newblock {PerceiverIO}: A general architecture for structured inputs \&
  outputs.
\newblock In {\em ICLR}, 2022.

\bibitem{kochanov2020kprnet}
Deyvid Kochanov, Fatemeh~Karimi Nejadasl, and Olaf Booij.
\newblock {KPRNet}: Improving projection-based {LiDAR} semantic segmentation.
\newblock In {\em ECCV}, 2020.

\bibitem{krahenbuhl2011efficient}
Philipp Kr{\"a}henb{\"u}hl and Vladlen Koltun.
\newblock Efficient inference in fully connected {CRF}s with gaussian edge
  potentials.
\newblock In {\em NeurIPS}, 2011.

\bibitem{landrieu2018large}
Loic Landrieu and Martin Simonovsky.
\newblock Large-scale point cloud semantic segmentation with superpoint graphs.
\newblock In {\em CVPR}, 2018.

\bibitem{li2022deepfusion}
Yingwei Li, Adams~Wei Yu, Tianjian Meng, Ben Caine, Jiquan Ngiam, Daiyi Peng,
  Junyang Shen, Yifeng Lu, Denny Zhou, Quoc~V Le, et~al.
\newblock {DeepFusion}: {LiDAR}-camera deep fusion for multi-modal {3D} object
  detection.
\newblock In {\em CVPR}, 2022.

\bibitem{lin2017focal}
Tsung-Yi Lin, Priya Goyal, Ross Girshick, Kaiming He, and Piotr Doll{\'a}r.
\newblock Focal loss for dense object detection.
\newblock In {\em ICCV}, 2017.

\bibitem{liu2022masked}
Haotian Liu, Mu Cai, and Yong~Jae Lee.
\newblock Masked discrimination for self-supervised learning on point clouds.
\newblock In {\em ECCV}, 2022.

\bibitem{liu2021learning}
Yueh-Cheng Liu, Yu-Kai Huang, Hung-Yueh Chiang, Hung-Ting Su, Zhe-Yu Liu,
  Chin-Tang Chen, Ching-Yu Tseng, and Winston~H Hsu.
\newblock Learning from {2D}: Contrastive pixel-to-point knowledge transfer for
  {3D} pretraining.
\newblock In {\em arXiv}, 2021.

\bibitem{loshchilov2016sgdr}
Ilya Loshchilov and Frank Hutter.
\newblock {SGDR}: Stochastic gradient descent with warm restarts.
\newblock In {\em ICLR}, 2017.

\bibitem{loshchilov2017decoupled}
Ilya Loshchilov and Frank Hutter.
\newblock Decoupled weight decay regularization.
\newblock In {\em ICLR}, 2019.

\bibitem{meng2019vv}
Hsien-Yu Meng, Lin Gao, Yu-Kun Lai, and Dinesh Manocha.
\newblock {VV-Net}: Voxel {VAE} net with group convolutions for point cloud
  segmentation.
\newblock In {\em ICCV}, 2019.

\bibitem{milioto2019rangenet++}
Andres Milioto, Ignacio Vizzo, Jens Behley, and Cyrill Stachniss.
\newblock {RangeNet++}: Fast and accurate {LiDAR} semantic segmentation.
\newblock In {\em IROS}, 2019.

\bibitem{pham2019jsis3d}
Quang-Hieu Pham, Thanh Nguyen, Binh-Son Hua, Gemma Roig, and Sai-Kit Yeung.
\newblock {JSIS3D}: Joint semantic-instance segmentation of {3D} point clouds
  with multi-task pointwise networks and multi-value conditional random fields.
\newblock In {\em CVPR}, 2019.

\bibitem{qi2017pointnet}
Charles~R Qi, Hao Su, Kaichun Mo, and Leonidas~J Guibas.
\newblock {PointNet}: Deep learning on point sets for {3D} classification and
  segmentation.
\newblock In {\em CVPR}, 2017.

\bibitem{qi2017pointnet++}
Charles~Ruizhongtai Qi, Li Yi, Hao Su, and Leonidas~J Guibas.
\newblock {PointNet++}: Deep hierarchical feature learning on point sets in a
  metric space.
\newblock In {\em NeurIPS}, 2017.

\bibitem{qian2022pix4point}
Guocheng Qian, Xingdi Zhang, Abdullah Hamdi, and Bernard Ghanem.
\newblock {Pix4Point}: Image pretrained transformers for {3D} point cloud
  understanding.
\newblock In {\em arXiv}, 2022.

\bibitem{radford2021learning}
Alec Radford, Jong~Wook Kim, Chris Hallacy, Aditya Ramesh, Gabriel Goh,
  Sandhini Agarwal, Girish Sastry, Amanda Askell, Pamela Mishkin, Jack Clark,
  et~al.
\newblock Learning transferable visual models from natural language
  supervision.
\newblock In {\em ICML}, 2021.

\bibitem{razani2021lite}
Ryan Razani, Ran Cheng, Ehsan Taghavi, and Liu Bingbing.
\newblock {Lite-HDSeg}: {LiDAR} semantic segmentation using lite harmonic dense
  convolutions.
\newblock In {\em ICRA}, 2021.

\bibitem{sautier2022image}
Corentin Sautier, Gilles Puy, Spyros Gidaris, Alexandre Boulch, Andrei Bursuc,
  and Renaud Marlet.
\newblock Image-to-{LiDAR} self-supervised distillation for autonomous driving
  data.
\newblock In {\em CVPR}, 2022.

\bibitem{shi2016real}
Wenzhe Shi, Jose Caballero, Ferenc Husz{\'a}r, Johannes Totz, Andrew~P. Aitken,
  Rob Bishop, Daniel Rueckert, and Zehan Wang.
\newblock Real-time single image and video super-resolution using an efficient
  sub-pixel convolutional neural network.
\newblock In {\em CVPR}, 2016.

\bibitem{strudel2021segmenter}
Robin Strudel, Ricardo Garcia, Ivan Laptev, and Cordelia Schmid.
\newblock Segmenter: Transformer for semantic segmentation.
\newblock In {\em ICCV}, 2021.

\bibitem{tchapmi2017segcloud}
Lyne Tchapmi, Christopher Choy, Iro Armeni, JunYoung Gwak, and Silvio Savarese.
\newblock {SEGCloud}: Semantic segmentation of {3D} point clouds.
\newblock In {\em 3DV}, 2017.

\bibitem{thomas2019kpconv}
Hugues Thomas, Charles~R Qi, Jean-Emmanuel Deschaud, Beatriz Marcotegui,
  Fran{\c{c}}ois Goulette, and Leonidas~J Guibas.
\newblock {KPConv}: Flexible and deformable convolution for point clouds.
\newblock In {\em CVPR}, 2019.

\bibitem{touvron2022three}
Hugo Touvron, Matthieu Cord, Alaaeldin El-Nouby, Jakob Verbeek, and Herv{\'e}
  J{\'e}gou.
\newblock Three things everyone should know about vision transformers.
\newblock In {\em ECCV}, 2022.

\bibitem{touvron2021going}
Hugo Touvron, Matthieu Cord, Alexandre Sablayrolles, Gabriel Synnaeve, and
  Herv{\'e} J{\'e}gou.
\newblock Going deeper with image transformers.
\newblock In {\em ICCV}, 2021.

\bibitem{triess2020scan}
Larissa~T Triess, David Peter, Christoph~B Rist, and J~Marius Z{\"o}llner.
\newblock Scan-based semantic segmentation of {LiDAR} point clouds: An
  experimental study.
\newblock In {\em IV}, 2020.

\bibitem{vaswani2017attention}
Ashish Vaswani, Noam Shazeer, Niki Parmar, Jakob Uszkoreit, Llion Jones,
  Aidan~N Gomez, {\L}ukasz Kaiser, and Illia Polosukhin.
\newblock Attention is all you need.
\newblock In {\em NeurIPS}, 2017.

\bibitem{velivckovic2017graph}
Petar Veli{\v{c}}kovi{\'c}, Guillem Cucurull, Arantxa Casanova, Adriana Romero,
  Pietro Lio, and Yoshua Bengio.
\newblock Graph attention networks.
\newblock In {\em ICLR}, 2018.

\bibitem{wang2021pointaugmenting}
Chunwei Wang, Chao Ma, Ming Zhu, and Xiaokang Yang.
\newblock {PointAugmenting}: Cross-modal augmentation for {3D} object
  detection.
\newblock In {\em CVPR}, 2021.

\bibitem{wang2019graph}
Lei Wang, Yuchun Huang, Yaolin Hou, Shenman Zhang, and Jie Shan.
\newblock Graph attention convolution for point cloud semantic segmentation.
\newblock In {\em CVPR}, 2019.

\bibitem{wang2022can}
Yi Wang, Zhiwen Fan, Tianlong Chen, Hehe Fan, and Zhangyang Wang.
\newblock Can we solve {3D} vision tasks starting from a {2D} vision
  transformer?
\newblock In {\em arXiv}, 2022.

\bibitem{wang2019dynamic}
Yue Wang, Yongbin Sun, Ziwei Liu, Sanjay~E Sarma, Michael~M Bronstein, and
  Justin~M Solomon.
\newblock Dynamic graph {CNN} for learning on point clouds.
\newblock In {\em TOG}, 2019.

\bibitem{wang2022bridged}
Yikai Wang, TengQi Ye, Lele Cao, Wenbing Huang, Fuchun Sun, Fengxiang He, and
  Dacheng Tao.
\newblock {Bridged Transformer} for vision and point cloud {3D} object
  detection.
\newblock In {\em CVPR}, 2022.

\bibitem{wu2019pointconv}
Wenxuan Wu, Zhongang Qi, and Li Fuxin.
\newblock {PointConv}: Deep convolutional networks on {3D} point clouds.
\newblock In {\em CVPR}, 2019.

\bibitem{xiao2021early}
Tete Xiao, Mannat Singh, Eric Mintun, Trevor Darrell, Piotr Doll{\'a}r, and
  Ross Girshick.
\newblock Early convolutions help transformers see better.
\newblock In {\em NeurIPS}, 2021.

\bibitem{xu2020squeezesegv3}
Chenfeng Xu, Bichen Wu, Zining Wang, Wei Zhan, Peter Vajda, Kurt Keutzer, and
  Masayoshi Tomizuka.
\newblock {SqueezeSegV3}: Spatially-adaptive convolution for efficient
  point-cloud segmentation.
\newblock In {\em ECCV}, 2020.

\bibitem{xu2021image2point}
Chenfeng Xu, Shijia Yang, Bohan Zhai, Bichen Wu, Xiangyu Yue, Wei Zhan, Peter
  Vajda, Kurt Keutzer, and Masayoshi Tomizuka.
\newblock {Image2Point}: {3D} point-cloud understanding with pretrained {2D}
  convnets.
\newblock In {\em arXiv}, 2021.

\bibitem{yan2020pointasnl}
Xu Yan, Chaoda Zheng, Zhen Li, Sheng Wang, and Shuguang Cui.
\newblock {PointASNL}: Robust point clouds processing using nonlocal neural
  networks with adaptive sampling.
\newblock In {\em CVPR}, 2020.

\bibitem{yu2022point}
Xumin Yu, Lulu Tang, Yongming Rao, Tiejun Huang, Jie Zhou, and Jiwen Lu.
\newblock {Point-BERT}: Pre-training {3D} point cloud transformers with masked
  point modeling.
\newblock In {\em CVPR}, 2022.

\bibitem{zhang2020deep}
Feihu Zhang, Jin Fang, Benjamin Wah, and Philip Torr.
\newblock Deep {FusionNet} for point cloud semantic segmentation.
\newblock In {\em ECCV}, 2020.

\bibitem{zhang2022cat}
Yanan Zhang, Jiaxin Chen, and Di Huang.
\newblock {CAT-Det}: Contrastively augmented transformer for multi-modal {3D}
  object detection.
\newblock In {\em CVPR}, 2022.

\bibitem{zhang2020polarnet}
Yang Zhang, Zixiang Zhou, Philip David, Xiangyu Yue, Zerong Xi, Boqing Gong,
  and Hassan Foroosh.
\newblock {PolarNet}: An improved grid representation for online {LiDAR} point
  clouds semantic segmentation.
\newblock In {\em CVPR}, 2020.

\bibitem{zhao2021point}
Hengshuang Zhao, Li Jiang, Jiaya Jia, Philip~HS Torr, and Vladlen Koltun.
\newblock {Point Transformer}.
\newblock In {\em ICCV}, 2021.

\bibitem{zhou2022self}
Junsheng Zhou, Xin Wen, Yu-Shen Liu, Yi Fang, and Zhizhong Han.
\newblock Self-supervised point cloud representation learning with occlusion
  auto-encoder.
\newblock In {\em arXiv}, 2022.

\bibitem{zhu2021cylindrical}
Xinge Zhu, Hui Zhou, Tai Wang, Fangzhou Hong, Wei Li, Yuexin Ma, Hongsheng Li,
  Ruigang Yang, and Dahua Lin.
\newblock Cylindrical and asymmetrical {3D} convolution networks for
  {LiDAR}-based perception.
\newblock In {\em CVPR}, 2021.

\bibitem{zhuang2021perception}
Zhuangwei Zhuang, Rong Li, Kui Jia, Qicheng Wang, Yuanqing Li, and Mingkui Tan.
\newblock Perception-aware multi-sensor fusion for {3D} {LiDAR} semantic
  segmentation.
\newblock In {\em ICCV}, 2021.

\end{thebibliography}
